\documentclass[10pt,twocolumn,letterpaper]{article}

\usepackage{cvpr}              

\usepackage[dvipsnames]{xcolor}
\newcommand{\red}[1]{{\color{red}#1}}

\let\oldthebibliography\thebibliography
\let\oldendthebibliography\endthebibliography
\renewenvironment{thebibliography}[1]{%
    \oldthebibliography{#1}%
    \setcounter{enumiv}{41}%
}{\oldendthebibliography}

\definecolor{cvprblue}{rgb}{0.21,0.49,0.74}
\usepackage[pagebackref,breaklinks,colorlinks,citecolor=cvprblue]{hyperref}
\usepackage{url}            
\usepackage{booktabs}       
\usepackage{amsfonts}       
\usepackage{nicefrac}       
\usepackage{microtype}      
\usepackage[dvipsnames]{xcolor}
\usepackage{graphicx}
\usepackage{float}
\usepackage{tabu}
\usepackage{rotating}
\usepackage{amssymb}
\usepackage{bm}
\usepackage{graphics}
\usepackage{multirow}
\usepackage{arydshln}
\usepackage{fixmath}
\usepackage{algorithm}
\usepackage{algpseudocode}
\usepackage{pifont}
\usepackage{mathtools}
\usepackage{xfrac}
\usepackage{arydshln}
\usepackage{multirow}
\usepackage[accsupp]{axessibility}
\newcommand{\minisection}[1]{\vspace{2mm}\noindent{\textbf{#1}.}}

\usepackage{amsmath}
\DeclareMathOperator{\argmax}{argmax} 

\usepackage[capitalize]{cleveref}
\crefname{section}{Sec.}{Secs.}
\Crefname{section}{Section}{Sections}
\Crefname{table}{Table}{Tables}
\crefname{table}{Tab.}{Tabs.}

\newcommand{\Eref}[1]{Eq.~(\ref{#1})}

\newcommand{\psnr}[1]{{\textcolor{RedViolet}{#1}}}
\newcommand{\ssim}[1]{{\textcolor{Blue}{#1}}}

\def\paperID{11140} 
\def\confName{CVPR}
\def\confYear{2024}

\title{Blur2Blur: Blur Conversion for Unsupervised Image Deblurring \\ on Unknown Domains}

\author{
Bang-Dang Pham$^{1}$ \quad Phong Tran$^{2}$ \quad Anh Tran$^{1}$ \quad Cuong Pham$^{1,3}$ \quad Rang Nguyen$^{1}$\quad Minh Hoai$^{1,4}$\\ 
\small{\textsuperscript{1}VinAI Research, Vietnam \quad \textsuperscript{2}MBZUAI, UAE \quad \textsuperscript{3}Posts \& Telecommunications Inst. of Tech., Vietnam
\quad \textsuperscript{4}University of Adelaide, Australia}\\
\texttt{\scriptsize \{v.dangpb1, v.anhtt152, v.rangnhm, v.hoainm\}@vinai.io} \quad 
\texttt{\scriptsize cuongpv@ptit.edu.vn} \quad
\texttt{\scriptsize the.tran@mbzuai.ac.ae} 
}
\begin{document}
\linespread{0.925}

\def\mA{\mathcal{A}}
\def\mB{\mathcal{B}}
\def\mC{\mathcal{C}}
\def\mD{\mathcal{D}}
\def\mE{\mathcal{E}}
\def\mF{\mathcal{F}}
\def\mG{\mathcal{G}}
\def\mH{\mathcal{H}}
\def\mI{\mathcal{I}}
\def\mJ{\mathcal{J}}
\def\mK{\mathcal{K}}
\def\mL{\mathcal{L}}
\def\mM{\mathcal{M}}
\def\mN{\mathcal{N}}
\def\mO{\mathcal{O}}
\def\mP{\mathcal{P}}
\def\mQ{\mathcal{Q}}
\def\mR{\mathcal{R}}
\def\mS{\mathcal{S}}
\def\mT{\mathcal{T}}
\def\mU{\mathcal{U}}
\def\mV{\mathcal{V}}
\def\mW{\mathcal{W}}
\def\mX{\mathcal{X}}
\def\mY{\mathcal{Y}}
\def\mZ{\mathcal{Z}} 

\def\bbN{\mathbb{N}} 
\def\bbR{\mathbb{R}} 
\def\bbP{\mathbb{P}} 
\def\bbQ{\mathbb{Q}} 
\def\bbE{\mathbb{E}}

\def\1n{\mathbf{1}_n}
\def\0{\mathbf{0}}
\def\1{\mathbf{1}}

\def\A{{\bf A}}
\def\B{{\bf B}}
\def\C{{\bf C}}
\def\D{{\bf D}}
\def\E{{\bf E}}
\def\F{{\bf F}}
\def\G{{\bf G}}
\def\H{{\bf H}}
\def\I{{\bf I}}
\def\J{{\bf J}}
\def\K{{\bf K}}
\def\L{{\bf L}}
\def\M{{\bf M}}
\def\N{{\bf N}}
\def\O{{\bf O}}
\def\P{{\bf P}}
\def\Q{{\bf Q}}
\def\R{{\bf R}}
\def\S{{\bf S}}
\def\T{{\bf T}}
\def\U{{\bf U}}
\def\V{{\bf V}}
\def\W{{\bf W}}
\def\X{{\bf X}}
\def\Y{{\bf Y}}
\def\Z{{\bf Z}}

\def\a{{\bf a}}
\def\b{{\bf b}}
\def\c{{\bf c}}
\def\d{{\bf d}}
\def\e{{\bf e}}
\def\f{{\bf f}}
\def\g{{\bf g}}
\def\h{{\bf h}}
\def\i{{\bf i}}
\def\j{{\bf j}}
\def\k{{\bf k}}
\def\l{{\bf l}}
\def\m{{\bf m}}
\def\n{{\bf n}}
\def\o{{\bf o}}
\def\p{{\bf p}}
\def\q{{\bf q}}
\def\r{{\bf r}}
\def\s{{\bf s}}
\def\t{{\bf t}}
\def\u{{\bf u}}
\def\v{{\bf v}}
\def\w{{\bf w}}
\def\x{{\bf x}}
\def\y{{\bf y}}
\def\z{{\bf z}}

\def\balpha{\mbox{\boldmath{$\alpha$}}}
\def\bbeta{\mbox{\boldmath{$\beta$}}}
\def\bdelta{\mbox{\boldmath{$\delta$}}}
\def\bgamma{\mbox{\boldmath{$\gamma$}}}
\def\blambda{\mbox{\boldmath{$\lambda$}}}
\def\bsigma{\mbox{\boldmath{$\sigma$}}}
\def\btheta{\mbox{\boldmath{$\theta$}}}
\def\bomega{\mbox{\boldmath{$\omega$}}}
\def\bxi{\mbox{\boldmath{$\xi$}}}
\def\bnu{\mbox{\boldmath{$\nu$}}}                                  
\def\bphi{\mbox{\boldmath{$\phi$}}}
\def\bmu{\mbox{\boldmath{$\mu$}}}

\def\bDelta{\mbox{\boldmath{$\Delta$}}}
\def\bOmega{\mbox{\boldmath{$\Omega$}}}
\def\bPhi{\mbox{\boldmath{$\Phi$}}}
\def\bLambda{\mbox{\boldmath{$\Lambda$}}}
\def\bSigma{\mbox{\boldmath{$\Sigma$}}}
\def\bGamma{\mbox{\boldmath{$\Gamma$}}}
                                  
\newcommand{\myprob}[1]{\mathop{\mathbb{P}}_{#1}}

\newcommand{\myexp}[1]{\mathop{\mathbb{E}}_{#1}}

\newcommand{\mydelta}[1]{1_{#1}}

\newcommand{\myminimum}[1]{\mathop{\textrm{minimum}}_{#1}}
\newcommand{\mymaximum}[1]{\mathop{\textrm{maximum}}_{#1}}    
\newcommand{\mymin}[1]{\mathop{\textrm{minimize}}_{#1}}
\newcommand{\mymax}[1]{\mathop{\textrm{maximize}}_{#1}}
\newcommand{\mymins}[1]{\mathop{\textrm{min.}}_{#1}}
\newcommand{\mymaxs}[1]{\mathop{\textrm{max.}}_{#1}}  
\newcommand{\myargmin}[1]{\mathop{\textrm{argmin}}_{#1}} 
\newcommand{\myargmax}[1]{\mathop{\textrm{argmax}}_{#1}} 
\newcommand{\myst}{\textrm{s.t. }}

\newcommand{\denselist}{\itemsep -1pt}
\newcommand{\sparselist}{\itemsep 1pt}

\definecolor{pink}{rgb}{0.9,0.5,0.5}
\definecolor{purple}{rgb}{0.5, 0.4, 0.8}   
\definecolor{gray}{rgb}{0.3, 0.3, 0.3}
\definecolor{mygreen}{rgb}{0.2, 0.6, 0.2}

\newcommand{\cyan}[1]{\textcolor{cyan}{#1}}
\newcommand{\blue}[1]{\textcolor{blue}{#1}}
\newcommand{\magenta}[1]{\textcolor{magenta}{#1}}
\newcommand{\pink}[1]{\textcolor{pink}{#1}}
\newcommand{\green}[1]{\textcolor{green}{#1}} 
\newcommand{\gray}[1]{\textcolor{gray}{#1}}    
\newcommand{\mygreen}[1]{\textcolor{mygreen}{#1}}    
\newcommand{\purple}[1]{\textcolor{purple}{#1}}       

\definecolor{greena}{rgb}{0.4, 0.5, 0.1}
\newcommand{\greena}[1]{\textcolor{greena}{#1}}

\definecolor{bluea}{rgb}{0, 0.4, 0.6}
\newcommand{\bluea}[1]{\textcolor{bluea}{#1}}
\definecolor{reda}{rgb}{0.6, 0.2, 0.1}
\newcommand{\reda}[1]{\textcolor{reda}{#1}}

\def\changemargin#1#2{\list{}{\rightmargin#2\leftmargin#1}\item[]}
\let\endchangemargin=\endlist
                                               
\newcommand{\cm}[1]{}

\newcommand{\mtodo}[1]{{\color{red}$\blacksquare$\textbf{[TODO: #1]}}}
\newcommand{\myheading}[1]{\vspace{1ex}\noindent \textbf{#1}}
\newcommand{\htimesw}[2]{\mbox{$#1$$\times$$#2$}}

\newcommand{\young}[1]{{\color{blue}$\blacksquare$\textbf{Alternative}: #1}}


\newif\ifshowsolution
\showsolutiontrue

\ifshowsolution  
\newcommand{\Solution}[2]{\paragraph{\bf $\bigstar $ SOLUTION:} {\sf #2} }
\newcommand{\Mistake}[2]{\paragraph{\bf $\blacksquare$ COMMON MISTAKE #1:} {\sf #2} \bigskip}
\else
\newcommand{\Solution}[2]{\vspace{#1}}
\fi

\newcommand{\truefalse}{
\begin{enumerate}
	\item True
	\item False
\end{enumerate}
}

\newcommand{\yesno}{
\begin{enumerate}
	\item Yes
	\item No
\end{enumerate}
}

\twocolumn[{%
\renewcommand\twocolumn[1][]{#1}%
\maketitle
\begin{center}
    \includegraphics[width=.95\linewidth]{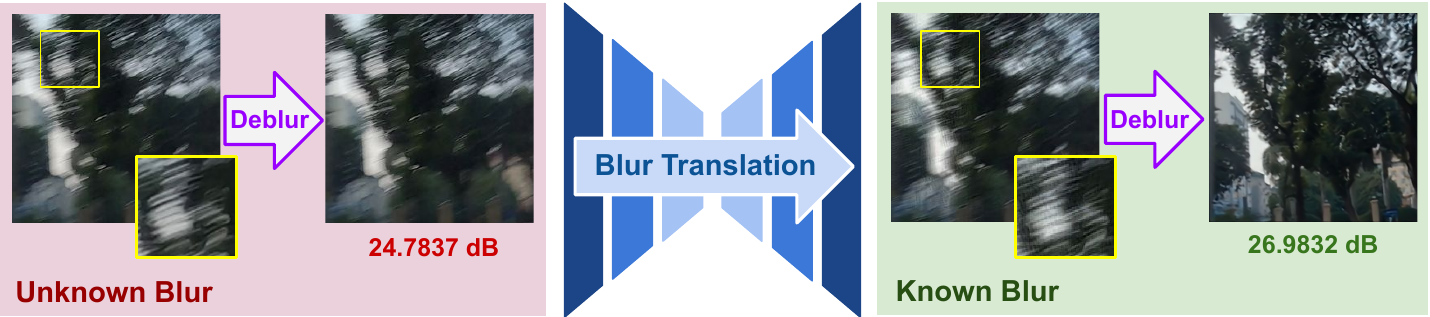}
    {\captionof{figure}{We address the unsupervised image deblurring problem by training a blur translator that converts an input image with unknown blur to an image with a predefined known blur. The figure shows the effectiveness of our approach. The blurry images before and after translation (left image in each box) exhibit similar visual content but have different blur patterns (zoomed-in patches). While a standard image deblurring technique fails to restore the unknown-blur image, it successfully recovers the known-blur version, yielding an approximate 2.2 dB increase in PSNR score (noted below each deblurred image on the right side of each box).}\label{fig:teaser}}
\end{center}%
}]
\maketitle

\begin{abstract}

This paper presents an innovative framework designed to train an image deblurring algorithm tailored to a specific camera device. This algorithm works by transforming a blurry input image, which is challenging to deblur, into another blurry image that is more amenable to deblurring. The transformation process, from one blurry state to another, leverages unpaired data consisting of sharp and blurry images captured by the target camera device. Learning this blur-to-blur transformation is inherently simpler than direct blur-to-sharp conversion, as it primarily involves modifying blur patterns rather than the intricate task of reconstructing fine image details. The efficacy of the proposed approach has been demonstrated through comprehensive experiments on various benchmarks, where it significantly outperforms state-of-the-art methods both quantitatively and qualitatively. Our code and data are available at  {\small \url{https://zero1778.github.io/blur2blur/}}

\end{abstract}    
\section{Introduction}
Motion blur in images and videos is a common issue, often resulting from camera shake or rapid movement within the scene. Such blur can detract from the aesthetic quality of the content and may undermine the performance of downstream computer vision applications. Consequently, an effective image deblurring method is essential in various contexts.

While the idea of deblurring images from arbitrary, diverse sources sounds impressive and broadly useful, the practical necessity, commercial value, and societal impact of image deblurring are frequently connected to specific application scenarios and particular cameras. For example, a mobile phone manufacturer might focus on integrating the most effective deblurring algorithm for the camera types used in their latest phone models. Similarly, a factory manager might consider installing ceiling-mounted cameras to identify errors on the assembly line, enhancing workforce efficiency. However, motion blur could significantly degrade the performance of computer vision algorithms meant to detect and track workers' hands and tools. In law enforcement, a police officer using a body-worn camera coupled with face recognition technology might find that motion blur hampers the accuracy of detecting faces and identifying fugitives. Therefore, in these scenarios, the development of a framework to customize a deblurring algorithm for specific cameras or camera types becomes crucial and represents a significant and growing need.

In this paper, we explore the question: How can we deblur images captured by specific cameras? Classical deblurring algorithms, which use signal processing or theoretical models of motion blur, are one option. Yet, their reliance on oversimplified blur models limits their effectiveness in addressing the complex motion blur encountered in real-world scenarios. An alternative is a data-driven approach that leverages advancements in machine learning. This approach involves using pre-trained deblurring networks developed through supervised learning, as illustrated by works such as \cite{chen2021hinet, chen2022simple, kupyn2018deblurgan, tao2018srndeblur, zamir2022restormer, suin2020spatially}. These networks, trained on extensive datasets of paired images, aim to transform blurry images into sharp ones. However, they often suffer from overfitting and tend to underperform on novel blurred images that were not captured by the cameras used to create their training datasets. Our empirical findings indicate that the performance of these models is still unsatisfactory when confronting unseen blurs produced by real-world cameras.

When pre-trained networks are unsuitable, the alternative is to develop a deblurring network specifically for our camera. However, this approach faces the challenge of not having access to paired training data, consisting of corresponding blurry and sharp images. Generating such data typically involves a sophisticated setup with a beam splitter, identical cameras operating at varying speeds, and capabilities for time synchronization, geometrical alignment, and color calibration \cite{rsblur, levin2009understanding, liang2020raw}. Often, the camera targeted for deblurring may not meet these stringent requirements, and arranging this setup is not feasible for many. Therefore, we are left with the option of utilizing unpaired data. Yet, training on unpaired data presents its own set of challenges due to the lack of supervision for restoring fine details that are missing or distorted in the blurry input images. Existing methods \cite{lu2019unsupervised, zhao2022fcl, zhu2017unpaired, yi2017dualgan}, which attempt to recreate these absent details, frequently fall short, particularly when dealing with blur typical of real-world images.

In this paper, we introduce \textbf{Blur2Blur}, a novel plug-and-play framework that leverages pretrained deblurring models to train an image deblurring algorithm specifically for a chosen camera device. Similar to other unsupervised deblurring methods, we utilize unpaired data. More precisely, we use the target camera to capture a set of blurry images and sharp images, without requiring a one-to-one correspondence between the images in these two sets. This approach makes data collection relatively simple and straightforward. Our method diverges from existing unsupervised methods by not attempting to directly learn a function from the domain of blurry images captured by our camera (the unknown blur domain, denoted as \(C\)), to the domain of sharp images. Instead, our strategy involves first learning a mapping \(G\) from the domain \(C\) to another domain \(C'\) of blurry images, where deblurring techniques are already well-established. To deblur an image taken by our camera, we first convert it into an image in \(C'\) using the learned mapping \(G\), then apply a pre-trained network to deblur this transformed image. Consequently, our primary goal is to learn the blur-to-blur mapping from \(C\) to \(C'\), which is inherently less challenging than the direct blur-to-sharp mapping, because the former primarily involves altering blur patterns rather than the more complex task of reconstructing detailed image features.

To learn the blur-to-blur mapping, we propose a novel learning framework to leverage the collected set of blurry and sharp images as well as the blurry images from the known blur domain $C'$. To train the blur-to-blur mapping network, we carefully define various loss terms, including perceptual, adversarial, and gradient penalty terms. The details of our approach are illustrated in \cref{fig:teaser}.

We conducted extensive experiments to compare the effectiveness of our model with other state-of-the-art image deblurring approaches on both real-world and synthetic blur datasets. The results demonstrate that Blur2Blur outperforms other methods by a significant margin, highlighting its superior performance in addressing the challenges of image deblurring in real-world settings. Notably, when combined with our blur translation method, supervised methods achieve an impressive boost up to 2.91 dB in PSNR.
\section{Related Work}

Many methods have been proposed for image deblurring. Beyond classical methods that do not necessitate training data, many contemporary approaches are grounded in machine learning. Learning-based methods can be broadly categorized based on their data requirements, whether it be paired, synthetic, or unpaired data. This section reviews representative works from these categories. 

\myheading{Classical Image Deblurring.}
Early deblurring methods assume that the blur operator is linear and uniform. In other words, the blur can be approximated by a single convolution operator: $y = x * k + \eta$, where $y$, $x$, $k$, and $\eta$ represent the blurry image, sharp image, blur kernel, and noise, respectively. 
Based on this assumption, given a blurry image $y$, the sharp image $x$ and the blur kernel $k$ can be obtained by maximizing the posterior distribution: $x^*, k^* = \argmax_{x, k} P(x, k|y)P(x)P(k)$.
Traditional methods primarily focus on finding prior distributions for either $x$ \cite{chan1998total,krishnan2009fast,hradivs2015convolutional,krishnan2011blind} or $k$ \cite{ren2020neural,levin2009understanding,liu2014blind}. However, these methods generalize poorly to real-world blurry images because blur kernels are often non-uniform and non-linear.

\myheading{Supervised learning with paired data.} 
Going beyond the assumptions of uniformity and linearity, several deep deblurring neural networks have been proposed \cite{tao2018scale,zamir2021multi,chen2022simple,kupyn2018deblurgan, kupyn2019deblurgan, mehri2021mprnet}, demonstrating promising results. These networks are typically trained on large-scale datasets containing pairs of blurry and sharp images. The distinguishing factors among these works primarily lie on their architectural designs. For instance, \citet{tao2018scale} introduced a multi-scale recurrent network architecture specifically tailored for image deblurring. Other methods \cite{gopro, mimounet} leveraged a coarse-to-fine strategy, utilizing multi-scale inputs to incrementally refine the deblurring process. \citet{kupyn2018deblurgan} was the first to incorporate GAN-based loss into the image deblurring framework, aiming to enhance the realism of deblurred images. Meanwhile, \citet{zamir2021multi} proposed a multi-stage framework that breaks down the image restoration task into smaller, more manageable stages. Lastly, \citet{chen2022simple} 
reduced the complexity both between and within network blocks based on UNet \cite{ronneberger2015u} architecture.

In supervised learning, training convolutional networks effectively requires extensive datasets comprising both sharp and blurry image pairs. Acquiring these datasets can be a complex and lengthy process, often necessitating advanced hardware and careful setup. Recent studies \cite{pham2023hypercut,rim2020real,zhong2020efficient} have introduced real-world deblurring datasets created using a dual-camera system, consisting of a high-speed and a low-speed camera, synchronized and aligned precisely with a time trigger and a beam splitter. This method ensures the collection of perfectly matched pairs of blurry and sharp images. Nonetheless, a limitation arises as deblurring networks trained on these specific datasets may become too tailored to the characteristics of the cameras used, resulting in reduced performance when applied to images from different cameras. Moreover, the dual-camera system is an advanced setup, requiring specific camera types that meet certain criteria, which means not all cameras are suitable for this purpose.

\myheading{Supervised learning with synthesized data}. One common approach for synthesizing blurry images is to average multiple consecutive sharp frames from a video sequence \cite{gopro, nah2019ntire}. Although this method mimics the way blurry images are captured, it has been demonstrated that models trained on these datasets often underperform when tested on real-world blurry images \cite{tran2021explore, rim2022realistic}. Recent studies have proposed more advanced techniques to synthesize deblurring datasets, aiming to improve the generalization of models trained on these datasets to unseen blur. For instance, \citet{zhang2021designing} created a synthesized dataset by combining multiple types of degradation operators initially developed for the super-resolution task. \citet{rim2022realistic} compared real and synthetic blurry images to design a more realistic blur synthesis pipeline. However, as demonstrated in \cref{sec:exp_result}, the degradation augmentation in \cite{zhang2021designing} significantly impairs the quality of input images, leading to distorted outputs. On the other hand, models trained on the synthesized deblurring dataset in \cite{rim2022realistic} exhibit signs of overfitting to the training data.

One promising direction was to leverage the known relationship between blurry and sharp image pairs from existing datasets \cite{tran2021explore}. This method involves capturing the blur distribution characteristic of each pair, which can then be applied to construct a synthesized blurred dataset. Inspired by the effectiveness of this strategy in capturing blur attributes from the known dataset, Blur2Blur adopts this approach. It is designed to discern and retain the blur kernel while selectively ignoring the camera-specific attributes of the target dataset.

\myheading{Unupservised learning with unpaired data}. 
Another approach to address the overfitting problem is through unpaired deblurring \cite{lu2019unsupervised, zhao2022fcl, zhu2017unpaired, yi2017dualgan}. Unlike supervised methods, these techniques do not require paired sharp and blurry images for training. However, they often face limitations, such as being domain-specific \cite{lu2019unsupervised}, or making low-level statistical assumptions about blur operators \cite{zhao2022fcl}, which may not be valid for real-world blurry images. To facilitate domain adaptation between blurred and sharp images, other methods \cite{zhu2017unpaired, yi2017dualgan} have been explored. However, these approaches struggle to bridge the gap between these domains effectively due to (1) the significant variation in the degree of blur across different images, which affects the perceived semantics of the objects within, and (2) the complex and unpredictable nature of real-world blur patterns, often contradicting the simplistic assumptions used in these models. Consequently, the challenge of achieving truly blind image deblurring remains unsolved.

Considering these limitations, our Blur2Blur approach is centered around the innovative idea of blur kernel transfer. This involves transforming the blur kernel from any particular camera into a familiar blur kernel from a dataset or camera that has a strong, pre-trained deblurring model. This method enables us to utilize the benefits of supervised techniques within an unsupervised framework, effectively tackling the challenge of deblurring images with a wide range of unknown blur distributions.
\section{Methodology}
\label{overview}
\subsection{Approach Overview}
We formulate a blurry image $y$ as a function of the corresponding sharp image $x$ through a blur operator $\mathcal{F}_C(\cdot, k)$, which is associated with a device-dependent blur domain $C$ and a blur kernel $k$:
\begin{align}
    y = \mathcal{F}_C(x, k) + \eta, \label{eq:blur}
\end{align}
where $\eta$ is a noise term. Our task is to find a deblurring function $\mathcal{G}^*_C$ that can recover the sharp image from the blurry input, i.e., $\mathcal{G}^*_C(y) = x$.

One strategy is to utilize an existing, pre-trained deblurring network to approximate the desired function $\mG^*_C$, and then use it for deblurring. However, this approach often leads to unsatisfactory results. The pre-trained network is generally trained on a dataset from a camera with a unique blur space $C'$, which is likely to be different from the blur space $C$ of our camera. In essence, this would mean approximating $\mG^*_C$ with $\mG^*_{C'}$, an approach that is not ideal due to the differences between $C$ and $C'$, resulting in suboptimal deblurring performance.

When a pre-trained network is not a good choice, our remaining option is to train a new deblurring network tailored to our camera. The obstacle here is that the specific blur space $C$ of our device is unknown, and we cannot rely on having paired training data of corresponding blurry and sharp images. Paired training data requires a complex hardware setup, involving a beam splitter, along with identical devices capturing at different speeds, and the capability for time synchronization, geometrical alignment, and color calibration. Not all camera devices meet these requirements, and setting up such a system is beyond the expertise of many. Consequently, our only feasible option is to use unpaired data. Fortunately, we can access the camera device to capture sets of blurry images $\mB$ and sharp images $\mS$, which are unpaired and do not necessitate correspondence between images in $\mB$ and images in $\mS$. Thus, gathering these datasets is relatively easy and straightforward. The downside, however, is that learning from unpaired data is challenging. The deblurring process, which transforms a blurred image $y$ into a sharp image $x$, typically requires an understanding of the blurring domain $C$. For unpaired data, this necessity poses a significant hurdle, especially in reconstructing fine details absent or distorted in the blurred input. Traditional deblurring networks \cite{lu2019unsupervised, zhao2022fcl, zhu2017unpaired, yi2017dualgan}, attempting to `hallucinate' these missing details, often produce unsatisfactory results, particularly with images affected by real-world blurring.

In this section, we introduce an innovative method to learn $\mG^*_{C}$. Rather than directly learning this function, which is extremely challenging, or roughly approximating it using a function learned for another blur domain $\mG^*_{C'}$, we treat $\mG^*_{C}$ as a composition of $\mG^*_{C'}$ and a translation function $G$, i.e., $\mG^*_C = \mG^*_{C'} \circ G$. Our goal then shifts to learning $G$ to bridge the gap between domains $C$ and $C'$. 

More specifically, our task is to learn a mapping function $G$ that maps each blurry input image $y$ defined in \Eref{eq:blur} to an image $y'$ with the same sharp visual representation $x$ but belongs to a known blur distribution $C'$:
\begin{align}
    G: y \rightarrow y', \textrm{where } y' = \mF_{C'}(x, k') + \eta'.
    \label{eq:G}
\end{align}

Our approach breaks a complex task into two manageable ones. One task requires deblurring from $C'$, which, while challenging, benefits from existing research. We can select a well-performing pre-trained network $\mG^*_{C'}$, which has been trained with supervised learning using paired data in its domain. The other task is to learn a translation from an unknown blur domain $C$ to a known domain $C'$. The difficulty of this task depends on the differences between $C$ and $C'$, yet it is surely easier than directly learning a mapping from $C$ to a sharp domain. This is because a blur-to-blur transformation primarily modifies the blur patterns, avoiding the need to reconstruct intricate image details. Moreover, we have the flexibility to choose the most appropriate $C'$ and $\mG^*_{C'}$ for our specific blur domain. This flexibility extends to the possibility of utilizing synthetic data, which allows for the generation of extensive datasets, ensuring that the deblurring network is thoroughly trained.

In the remaining of this section, we will discuss two main components of our method, including the blur-to-blur translation network $G$ and the target blur space $C'$.

\begin{figure*}[t]
\centering
\begin{minipage}[b]{.41\linewidth}
\centering
    \includegraphics[width=\textwidth]{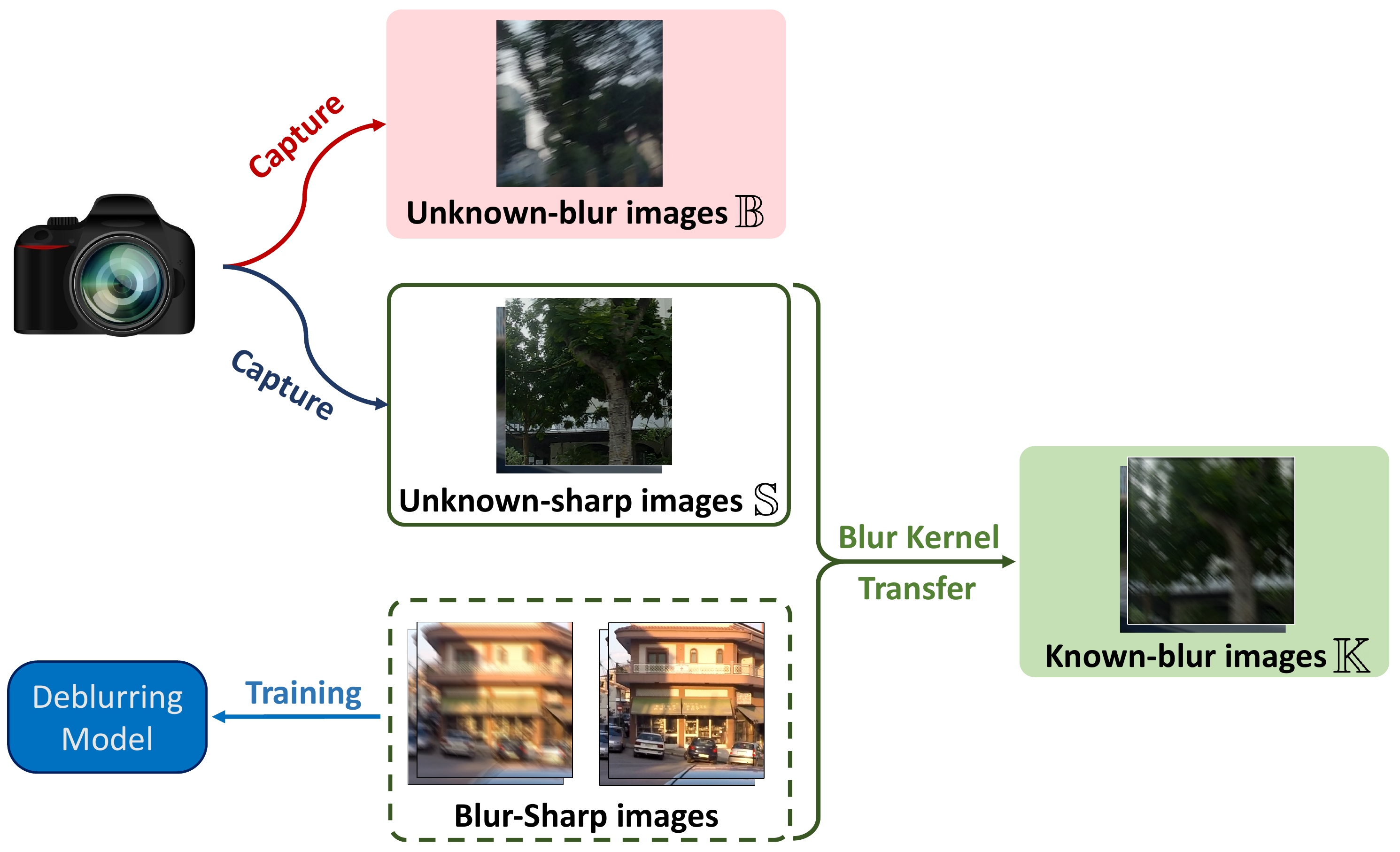}
    \subcaption{Unknown- and known-blur datasets}\label{fig:overview}
\end{minipage}
\hfill
\begin{minipage}[b]{.58\linewidth}
\centering
    \includegraphics[width=\textwidth]{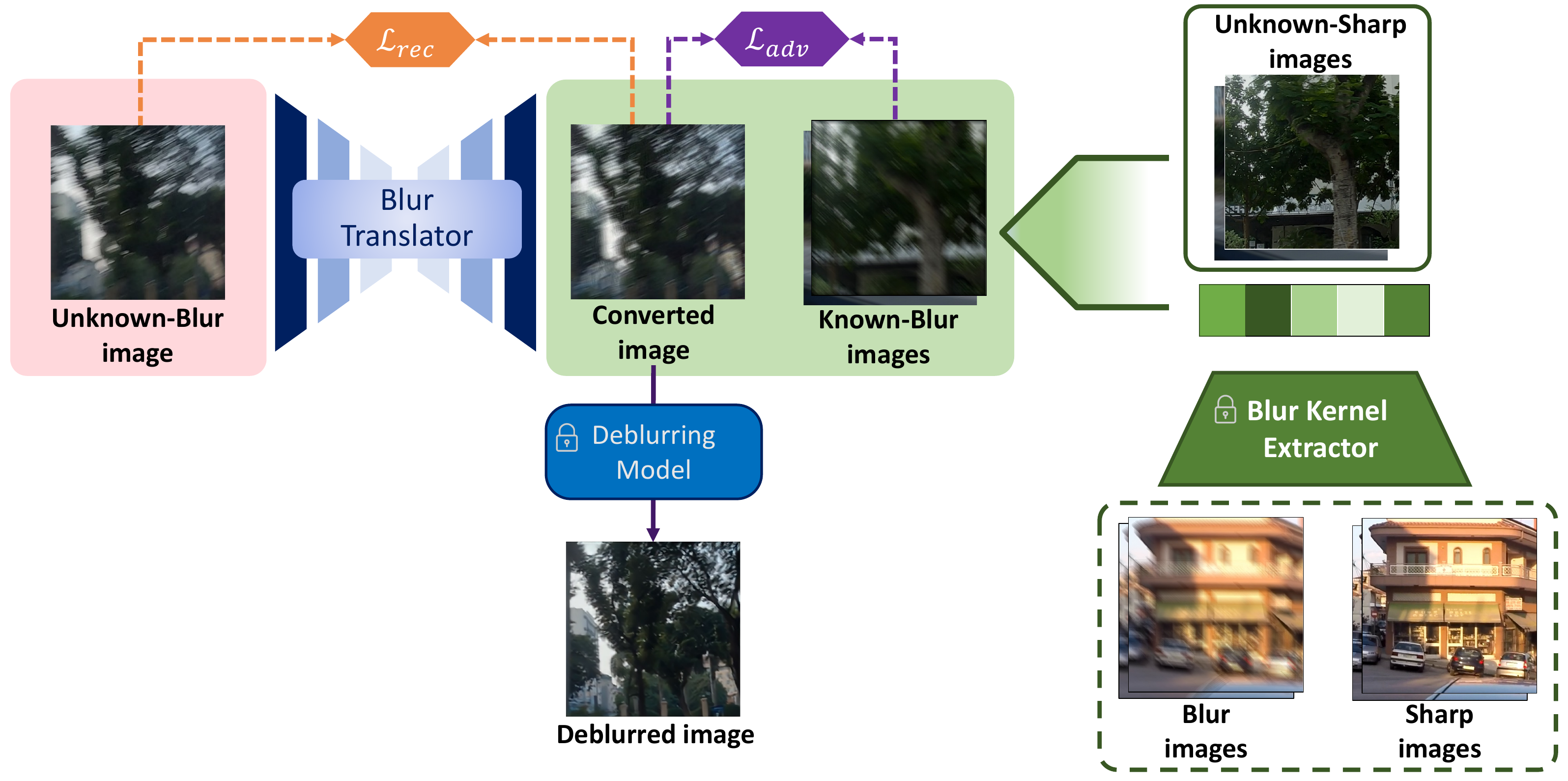}
    \subcaption{Blur translation}\label{fig:generator}
\end{minipage}
\vspace{-3mm}
\caption{\textbf{Overview of our problem and proposed method.} a) Given a camera, we aim to develop an algorithm to deblur its captured blurry images. We assume access to the camera to collect \textit{unpaired} sets of blurry images ($\mB$) and sharp image sequences ($\mS$). b) The key component in our proposed system is a blur translator that converts unknown-blur images captured by the camera to have the target known-blur presented in $\mK$. This translator is trained using reconstruction and adversarial losses. The converted images have known blur and can be successfully deblurred using the previously trained deblurring model (Zoom for best view).}
\vspace{-4mm}
\label{fig:system}
\end{figure*}

\subsection{Blur-to-blur translation}\label{sec:generator}

Our objective here is to train a blur-to-blur translation network $G$, capable of converting any blurry image from the unknown blur domain $C$ to a known blur domain $C'$ while preserving the image content. To train $G$, we require two datasets: $\mB$, which consists of blurry images from the unknown blur domain, and $\mK$ contains images with known blur, for which a deblurring model has already been trained. We design $G$ to work at multiple scales and carefully design the training losses to achieve the desired outcome.

\minisection{Adversarial Loss}
We employ an adversarial loss~\cite{goodfellow2020generative} to enforce the translation network $G$ to produce images with the desired target blur. To achieve this, we introduce a discriminator network $D$, which is responsible for distinguishing between real images from the known blur domain and generated images. Two networks $G$ and $D$ are trained alternately in a minimax game. The adversarial loss is defined as:
\begin{equation}
\begin{split}
\mathcal{L}_{adv}(G, D) = \; & \mathbb{E}_{y \sim \mK}[\log D(y)] \\
& + \mathbb{E}_{y \sim \mB}[\log (1 - D(G(y)))].
\end{split}
\end{equation}
The blur translation network is trained to minimize the above loss term, while the discriminator $D$ is trained to maximize it.
We also force the Lipschitz continuity constraint on the discriminator using the gradient penalty regularization \cite{gulrajani2017improved}:
\begin{equation}
\mathcal{L}^D_{grad}(D) = \mathbb{E}_{\hat{y} \sim \hat{\mB}}[(\|\nabla_{\hat{y}} D(\hat{y})\|_2 - 1)^2],
\end{equation}
where $\hat{\mB}$ is the set of samples $\hat{y}$ randomly interpolated between a real image $y \in \mB$ and the generated image $G(y)$ using a random mixing ratio $\epsilon \in [0, 1]$, i.e., $\hat{y} = \epsilon y + (1 - \epsilon) G(y)$.

\minisection{Reconstruction Loss}
Given a blurry image $y$, the desired function $G$ should translate the blur characteristics from $C$ to $C'$ while maintaining the elements belonging to the sharp image $x$. Using the adversarial loss helps translate the image to the target blur domain but does not guarantee sharp content preservation. Hence, we integrate a reconstruction loss to enforce the visual consistency between the generated blurry image $G(y)$ and the original image $y$. This loss term has two benefits: (1) it prevents $G$ from modifying the image content and only focuses on the blur kernel translation, and (2) it provides additional supervision to our network, enhancing the training stability. Moreover, to make $G$ focus on preserving the input semantic content rather than being overly constrained by pixel-wise accuracy, we (1) employ perceptual loss \cite{johnson2016perceptual} instead of the common $L_1$ or $L_2$ loss function and (2) adopt a multi-scale deblurring architecture \cite{mimounet} to reconstruct the image content from coarse to fine:
\begin{equation}
\mathcal{L}^G_{rec}(G) = \frac{1}{M} \sum_{i=1}^M \frac{1}{t_i}\mathbb{E}_{y_i \sim \mB}[||\phi(y_i) - \phi(G(y_i))||_1],
\end{equation}
where $M$ is the number of levels, $y_i$ is the input image at scale level $i$, $\phi(.)$ is a pre-trained feature extractor with the VGG19 backbone \cite{simonyan2015very}. We divide the loss by the number of total elements $t_i$ for normalization.

\minisection{Total Loss}
Our final objective function for $G$ combines the adversarial and reconstruction loss terms:
\begin{equation}
\mathcal{L}^G_{total}(G, D) = \mathcal{L}_{adv}(G, D) + \lambda_{rec}\mathcal{L}_{rec}(G),
\end{equation}
where $\lambda_{rec}$ is the weight factor for the reconstruction loss, ensuring the input content is maintained. Concurrently, the objective function for $D$ is established as follows:
\begin{equation}
\begin{split}
\mathcal{L}^D_{total}(G, D) =  - \mathcal{L}_{adv}(G, D)  + \lambda_{grad}\mathcal{L}_{grad}(D).
\end{split}
\end{equation}
Here $\lambda_{grad}$ is a hyperparameter that controls the importance of the gradient penalty loss component.

\subsection{Known Blur Selection}\label{sec:knownblur}

The choice of $C'$ and its representative dataset $\mK$ is important because the difficulty of learning the blur translation network depends on the discrepancy between the two blur domains.
As described in \cref{sec:generator}, the representative dataset $\mK$ only affects the adversarial training losses. The translation network $G$ aims to convert images in $\mB$ to have similar blur characteristics as images in $\mK$ so that the discriminator $D$ cannot differentiate between the generated images and the real images in $\mK$. However, if $\mK$ and $\mB$ have different characteristics besides the blur kernel distribution, such as color tone, image resolution, or device-dependent noise pattern, $D$ may rely on them to differentiate real and generated images. It can cause $G$ to either fail to converge or introduce undesired characteristics from the representative dataset $\mK$ into the transferred outcomes.  

To avoid this issue, we propose generating images in $\mK$ from a set of sharp images $\mS$ captured with the same camera as $\mB$, thus sharing identical characteristics. These images are then augmented by blur kernels from a known domain, characterized by a dataset of blurry-sharp image pairs using the blur transfer technique~\cite{tran2021explore}. The blurry-sharp image pair dataset can be selected from commonly used image deblurring datasets like REDS \cite{nah2019ntire}, GOPRO \cite{gopro}, RSBlur \cite{rsblur}, and RB2V \cite{pham2023hypercut}, and we can utilize any deblurring network pre-trained on that dataset. A key component in \cite{tran2021explore} is a Blur Kernel Extractor $F$ that can isolate and transfer blur kernels from random blurry-sharp image pairs to the target sharp inputs. After applying this blur synthesis procedure, we obtain a known-blur image set $\mK$ that carries blur kernels from the known-blur domain while maintaining other camera-based characteristics similar to the unknown-blur images in $\mB$. Consequently, the discriminator can focus on distinguishing based on blur kernels, facilitating effective blur-to-blur translation training. The overview problem and pipeline of our method is illustrated in \cref{fig:system}.

\section{Experiments}

\subsection{Experimental Setups}
\subsubsection{Datasets and implementation details}
We evaluate our proposed method on four datasets. \textbf{REDS dataset \cite{nah2019ntire}} consists of 300 high-speed videos used to create synthetic blur. By ramping up the frame rate from 120 to 1920 fps and averaging frames with an inverse Camera Response Function (CRF), it simulates more realistic motion blur, differentiating it from other synthetic datasets \cite{shen2019human,nah2017deep}. \textbf{GoPro dataset \cite{gopro}} comprises 3,142 paired frames of sharp and blurred images, recorded at 240 frames per second. It employed a synthesis method akin to that of the REDS dataset but with a different camera response function. We utilize this dataset as the main target data for evaluating deblurring methods in combination with Blur2Blur. \textbf{RSBlur dataset \cite{rsblur}} contains 13,358 real blurred images. It provides sequences of sharp images alongside blurred ones for in-depth blur analysis and offers the higher resolution than similar datasets. Noise levels are also estimated to assess and compare to the noise present in real-world blur scenarios.  In this paper, for the evaluation on the publicly available RSBlur dataset, we utilized the official dataset alongside its Additional set at a lower sampling rate, rather than generating blurry images from the \textit{RSBlur\_sharps} set using dense sampling as described in the original paper.  \textbf{RB2V dataset \cite{pham2023hypercut}} comprises about 11,000 real-world pairs of a blurry image and a sharp image sequence for street categories, denoted as \textit{RB2V\_street}. 
Experiments on this dataset are crucial for confirming the effectiveness of our algorithm in handling real-world, camera-specific data.

\myheading{Train and test data}.
To address practical deblurring problems, our method assumes access to unpaired sets containing blurry images $\mB$ and sharp images $\mS$. When selecting a dataset as the source for our deblurring evaluation, we divide its training data into two disjoint subsets that capture different scenes with a specific ratio of 0.6:0.4. In the first subset, we select blurry images to form the unknown-blur image set $\mB$, while in the second subset, we choose sharp images to construct the sharp set $\mS$. For the chosen target dataset, representing the domain for blur kernel translation via the Blur2Blur mechanism, we employ the entire training dataset to train our Blur Kernel Extractor \cite{tran2021explore} and subsequently apply this extractor to map captured blur embeddings onto the sharp image set $\mS$, creating the known-blur image set $\mK$. The blurry images in the test data of the source dataset are used to evaluate image deblurring algorithms. The statistics of source image sets are reported in \cref{tab:dataset}.
\begin{table}[t]
    \centering
    \begin{tabular}{lccc}
         \multirow{2.5}{*}{Dataset}& \multicolumn{3}{c}{Number of data samples} \\
         \cmidrule(lr){2-4} \cmidrule(lr){2-4}
          & U. blur ($\mB$) & U. sharp ($\mS$) & \hspace*{0mm} Test \hspace*{0mm} \\
         \hline 
         RB2V\_Street & 5400 & 3600 & 2053 \\ 
         REDS   & 14400 & 9600 & 3000 \\
         RSBlur   & 8115 & 5410 & 8301 \\
         GoPro   & 1261 & 842 & 1111 \\
         \bottomrule
    \end{tabular}
    \vskip -0.1in 
    \caption{Statistics of datasets used as unknown domains.}    
    \label{tab:dataset}
    \vspace{-3mm}
\end{table}

\begin{figure*}[!htb]
    \centering
    \includegraphics[width=.99\linewidth]{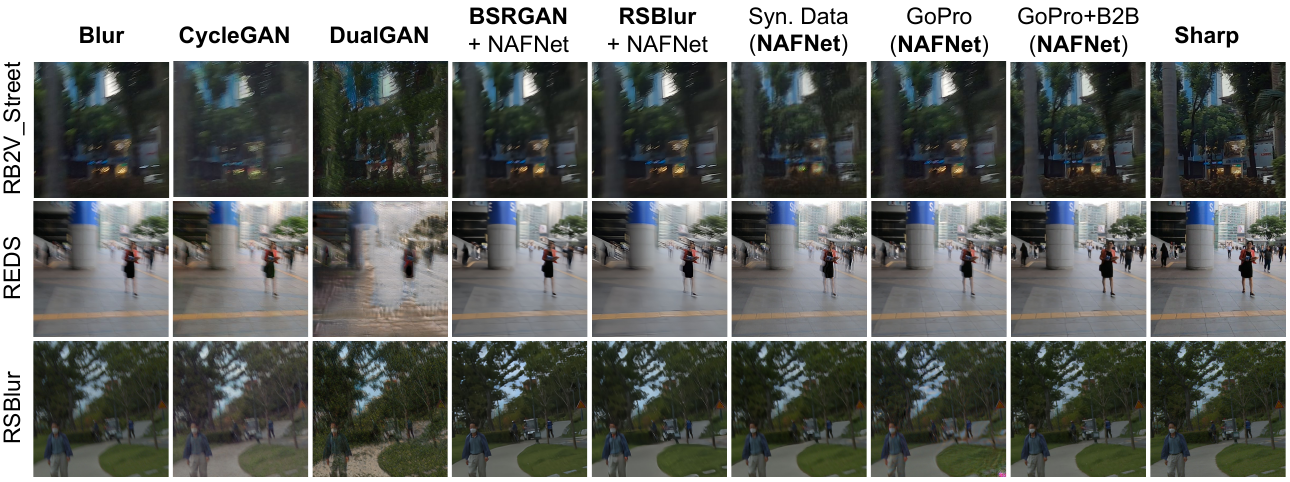}
    \vspace{-2mm}
    \caption{Comparing image deblurring results on three benchmark datasets with NAFNet. Due to
space limit, we skip the results with Restormer backbone, which is similar but slightly worse than
those with NAFNet. Best viewed when magnified on a digital display.}
    \vspace{-4mm}
    \label{fig:qual1}
\end{figure*}

\myheading {Implementation Details}.
We implemented the blur-to-blur translation network $G$ using MIMO-UNet \cite{mimounet} with the default configuration in Pix2Pix  \cite{isola2017image} implementation. For all experiments, we set the hyper-parameters $\lambda_{rec} = 0.8, \lambda_{grad}= 0.005$ and batch size of 16. To enhance our understanding of the network $G$ during its initial iterations, we sorted images based on their blur degree, determined by the variance of the responses obtained from applying the Laplacian operator. A lower variance corresponds to a reduced range of intensity changes, signaling a blurrier image with fewer edges. Initially, we optimized approximately 50\% of the data within a single batch. Subsequently, after 200K iterations, we incrementally scaled this proportion to encompass the full batch. We evaluated Blur2Blur in combination with different state-of-the-arts deblurring network backbones, including NAFNet \cite{chen2022simple} and Restormer \cite{zamir2022restormer}. During training, we randomly cropped these images to obtain a square shape of $256 {\times} 256$ and augmented with rotation, flip and color-jitter.

All experiments were performed using the Adam optimizer \cite{Kingma2014AdamAM}. Training our model required roughly 3 days for 1M iterations on 2 Nvidia A100 GPUs. 
The learning rate is maintained constant for the first 500K iterations and then linearly reduced during the remaining iterations as in \cite{torbunov2023rethinking}.

\subsubsection{Baselines}
We compared Blur2Blur with a comprehensive list of baseline methods from three categories: supervised methods (NAFNet \cite{chen2022simple}, Restormer \cite{zamir2022restormer}), unpaired training (CycleGAN \cite{zhu2017unpaired}, DualGAN \cite{yi2017dualgan}), and generalized image deblurring (BSRGAN+NAFNet \cite{zhang2021designing}, RSBlur+NAFNet \cite{rim2022realistic}). 

For fair comparisons, we retrained the supervised models using the blur-sharp pairs from the source dataset. Furthermore, to replicate real-world scenarios with the absence of paired data for deblurring network training, we generated synthetic motion-blur data derived from the unknown-sharp image set $\mS$ by adding motion blur synthesis techniques, such as the one provided by the \textit{imgaug} library \cite{imgaug}. This approach synthesized motion blur independently on each image.  For the unpaired training and generalized image deblurring approaches, we used the blurry images in $\mB$ and the sharp images from $\mS$ for training the deblurring network. BSRGAN was originally designed for blind image super-resolution, and we adapted it to work on blind image deblurring by adding motion blur augmentation (via averaging with neighboring frames) into its augmentation pipeline.

\subsection{Image Deblurring Results}\label{sec:exp_result}
To evaluate the performance of the Blur2Blur mechanism, we defined three data configurations. Each configuration consists of the Known-Blur dataset $\mK$, sourced from the GoPro dataset, and two unpaired datasets, $\mB$ and $\mS$, derived from the training partitions of the deblurring dataset REDS, RSBlur, or RB2V\_Street. For a comprehensive evaluation of the Blur2Blur model, we integrated it with two supervised image deblurring backbones, Restormer and NAFNet. Additionally, we compared the results with state-of-the-art baselines. The quantitative results are summarized in \cref{tab:quan}.

As observed, both unsupervised image deblurring and generalized deblurring approaches, despite expecting generalization power, exhibit poor performance on these challenging real-world datasets. Their scores are similar to, and sometimes significantly lower than, state-of-the-art supervised methods such as Restormer and NAFNet. In contrast, Blur2Blur demonstrates remarkable deblurring results. When combined with Restormer, Blur2Blur helps to increase the PSNR score by 2.63 dB on RB2V\_Street, 2.12 dB on REDS, and 2.91 dB on RSBlur. When combined with NAFNet, it provides consistent score increases, with 2.20 dB on RB2V\_Street, 2.31 dB on REDS, and 2.67 dB on RSBlur. 
NAFNet outperforms Restormer overall, making the combination of NAFNet and Blur2Blur the most effective approach. Moreover, our method comes close to matching the best results of supervised models trained on source datasets. 

We provide a qualitative comparison between image deblurring results in \cref{fig:qual1}. The comparison highlights a significant performance disparity between supervised methods and their counterparts. Unsupervised methods like DualGAN and CycleGAN struggle notably in deblurring, with DualGAN particularly unable to navigate the blur-to-sharp domain, tending instead to bridge the content and color distribution gap between the blurry ($\mB$) and sharp ($\mS$) datasets. Synthesis-based methods such as BSRGAN and RSBlur also fall short, failing to address unseen blurs, indicating the limitations of augmentation strategies, including those using the \textit{imgaug} library. Supervised method NAFNet  fails to handle unseen blurs, often yielding output mostly identical to the blurred inputs. However, our method effectively transforms unknown blurs into known ones. Our translation process successfully focuses on the blur kernel, minimizing bias from other image characteristics. By integrating Blur2Blur with NAFNet, we achieve a substantial recovery of high-quality sharp images, demonstrating the practical strength of our approach. Additional qualitative results for Restormer are provided in the supplementary material. 

\begin{table}[t]
\centering
\setlength{\tabcolsep}{3pt}
\resizebox{\columnwidth}{!}{%
\begin{tabular}{lccc}
\toprule  
& \textbf{RB2V\_Street} & \textbf{REDS} & \textbf{RSBlur} \\
\midrule
\textbf{NAFNet} \cite{chen2022simple} & & & \\
\quad w/ GoPro & \psnr{24.78} / \ssim{0.714} & \psnr{25.80} / \ssim{0.880} & \psnr{26.33} / \ssim{0.790} \\
\quad w/ Synthetic Data & \psnr{22.10} / \ssim{0.644} & \psnr{25.07} / \ssim{0.853} &\psnr{23.53} / \ssim{0.659} \\
\quad w/ Blur2Blur (GoPro) & \textbf{\psnr{26.98}} / \textbf{\ssim{0.812}} & \textbf{\psnr{28.11}} / \textbf{\ssim{0.893}} & \textbf{\psnr{29.00}} / \textbf{\ssim{0.857}} \\
\quad w/ \textit{the source domain*} & \psnr{28.72} / \ssim{0.883} & \psnr{29.09} / \ssim{0.927} & \psnr{33.06} / \ssim{0.888}\\
\textbf{Restormer} \cite{zamir2022restormer} & &  & \\
\quad w/ GoPro & \psnr{23.34} / \ssim{0.698} & \psnr{25.43} / \ssim{0.775} & \psnr{25.98} / \ssim{0.788} \\
\quad w/ Synthetic Data & \psnr{23.78} / \ssim{0.655} & \psnr{24.76} / \ssim{0.753} & \psnr{23.34} / \ssim{0.651} \\
\quad w/ Blur2Blur (GoPro) & \psnr{\underline{25.97}} / \ssim{\underline{0.750}} & \psnr{\underline{27.55}} / 
\ssim{\underline{0.885}} & \psnr{\underline{28.89}} / \ssim{\underline{0.850}} \\

\quad w/ \textit{the source domain*} & \psnr{27.43} / \ssim{0.849} &\psnr{28.23} / \ssim{0.916} & \psnr{32.87} / \ssim{0.874} \\
\midrule 
\textbf{Generalized Deblurring} & & & \\
BSRGAN \cite{zhang2021designing} & \psnr{23.31} / \ssim{0.645} & \psnr{26.39} / \ssim{0.803} & \psnr{27.11} / \ssim{0.810} \\
RSBlur \cite{rim2022realistic} & \psnr{23.42} / \ssim{0.603} & \psnr{26.32} / \ssim{0.812} & \psnr{26.98} / \ssim{0.798} \\
\midrule
\textbf{Unpaired Training} & & & \\
CycleGAN \cite{zhu2017unpaired} & \psnr{21.21} / \ssim{0.582} & \psnr{23.92} / \ssim{0.775} & \psnr{23.34} / \ssim{0.782} \\
DualGAN \cite{yi2017dualgan} & \psnr{21.02} / \ssim{0.556} & \psnr{23.50} / \ssim{0.700} & \psnr{22.78} / \ssim{0.704} \\
\bottomrule
\end{tabular} %
}
    \vskip -0.08in
\caption{Comparison of different deblurring methods on various datasets. For each test, we report \psnr{PSNR$\uparrow$} and \ssim{SSIM$\uparrow$} scores as evaluation metrics. The best scores are in \textbf{bold} and the second best score are in \underline{underline}. For a supervised method, NAFNet or Restormer, we assess its upper-bound of deblurring performance by training it on the \textit{training set of the source dataset*}.}
\vspace{-1mm}
\label{tab:quan}
\end{table}

\begin{table}[t] {
    \centering
    \setlength{\tabcolsep}{3pt}
    \begin{tabu}{l|ccccc}
        \toprule
        Ratio $\mB:\mS$ & 5:5 & 6:4 & 7:3 & 8:2 & 9:1 \\
        \midrule
        GoPro--RB2V\_Street & 26.02 & 26.98 & 26.92 & 25.98 & 24.32 \\
        GoPro--REDS & 27.53 & 28.11 & 28.10 & 27.00 & 26.43 \\
        \bottomrule
    \end{tabu}
    \vskip -0.1in
     \caption{PSNR debluring results with different Blur-to-Sharp ratios. \label{tab:ratio}}
    \vspace{-8mm}
}

\end{table}
\subsection{Blur2Blur Visualization}
\begin{figure}[t]
    \centering
    \includegraphics[width=.99\columnwidth]{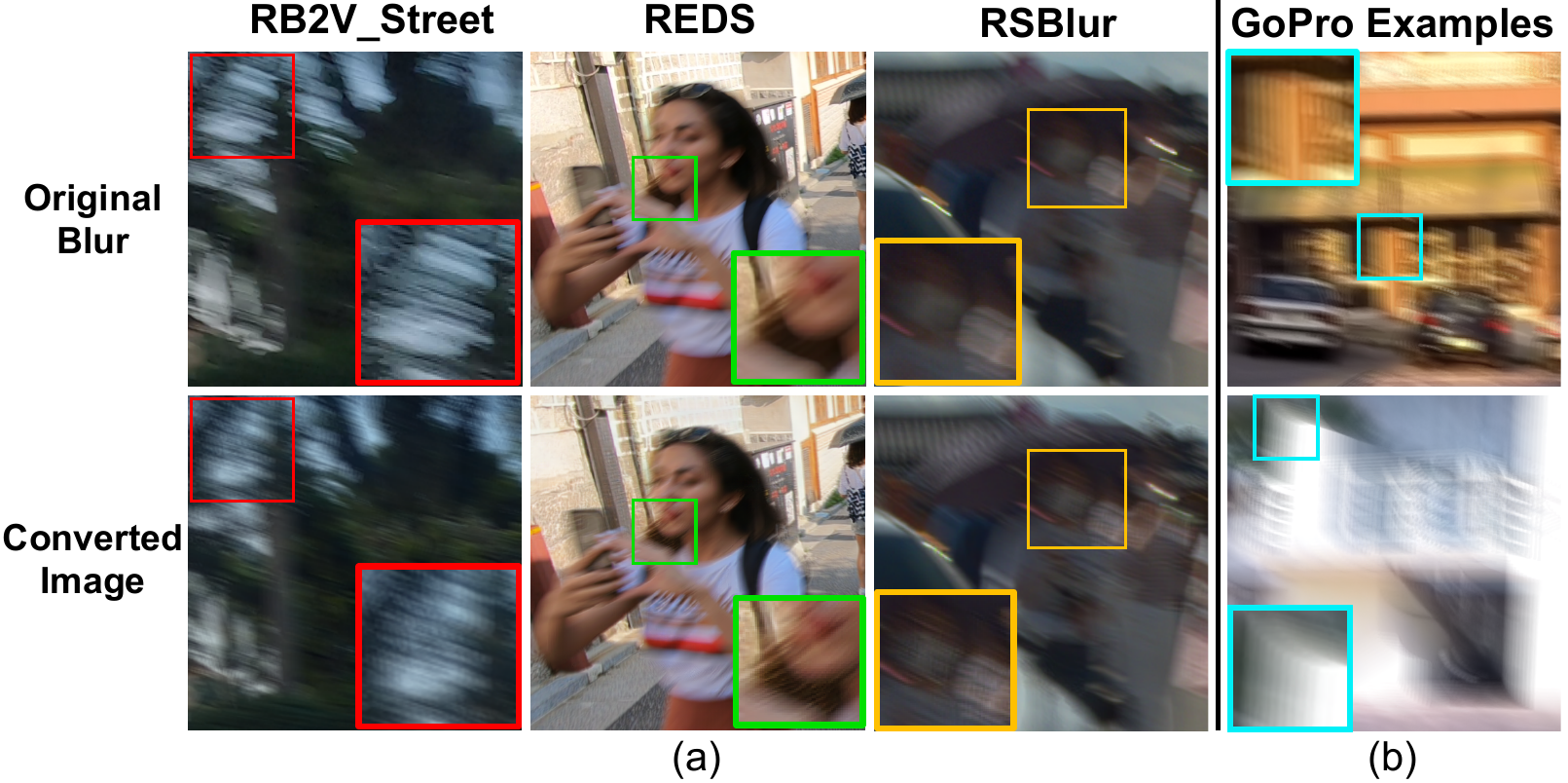}
    \vspace{-3mm}
    \caption{(a) A comparison of original images and their corresponding Blur2Blur converted version; (b) Selected examples demonstrating the GoPro dataset's blur pattern (Zoom for best view).}
    \vspace{-3mm}
    \label{fig:converted}
\end{figure}

\cref{fig:converted}\textcolor{red}{a} provides a comparative visualization between the original blurry image and its Blur2Blur converted images using the same source and target datasets as detailed in \cref{sec:exp_result}. As can be seen, our transformed images effectively adopt the blur pattern of the GoPro dataset, noted for its low sampling rate blur (as further shown in \cref{fig:converted}\textcolor{red}{b}), while preserving other content elements identical to the input. This demonstrates the Blur2Blur conversion's capability to 
produce transformed images that faithfully reflect the specific blur pattern while preserving the original content details.

\subsection{Ablation Study for Blur to Sharp Ratio} \label{sec:exp_abl}

We evaluate the significance of the blur-to-sharp ratio, represented as the ratio between datasets Unknown-Blur ($\mB$) and Unknown-Sharp ($\mS$). Specifically, we consider NAFNet as the image deblurring backbone, and consider the GoPro-RB2V\_Street and GoPro-REDS dataset settings, where GoPro represents our target camera device for which we have tailored a deblurring model. We conducted B2B experiments across a range of ratios from 5:5 to 9:1. The deblurring result in \cref{tab:ratio} demonstrates that a greater proportion of blurry images in the dataset, as seen in the 6:4 and 7:3 ratios, allows for a deeper understanding of the blur patterns characteristic of the target device, leading to
improved deblurring performance. However, excessively few sharp images, as in the 9:1 ratio, may cause the Blur2Blur method to overfit to limited sharp content.
To balance learning and prevent overfitting, a 6:4 ratio has been selected for all experiments in this study.

\begin{figure}[t]
    \centering
    \includegraphics[width=0.98\linewidth]{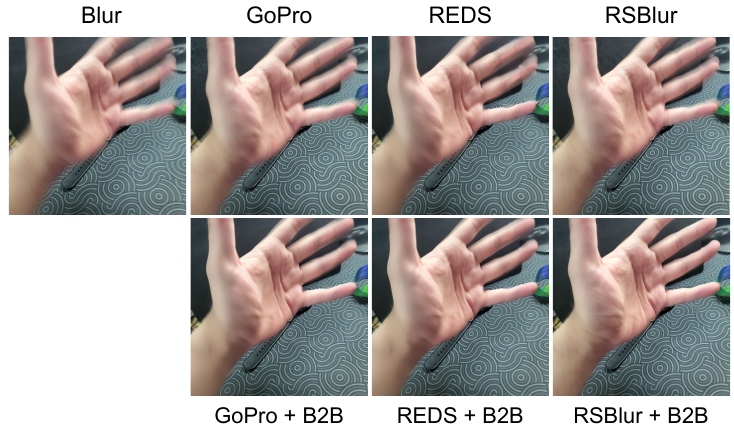}
    \vskip -0.1in 
    \caption{Qualitative comparison of deblurring models on the PhoneCraft dataset with multiple target datasets. \label{fig:qual_phonecraft}}    
    \vspace{-5mm}
\end{figure}

\subsection{Practicality Evaluation}
We evaluate the practicality of Blur2Blur in two imagined yet realistic scenarios. The first scenario involved a user desiring a deblurring algorithm for images taken with their smartphone camera. To facilitate this, we compiled a dataset named \textbf{PhoneCraft}, featuring images captured using a Samsung Galaxy Note 10 Plus. This dataset includes videos with motion-induced blur, refined through post-processing to remove other blur types, and clear, sharp videos recorded at 60fps. Over two hours, a variety of scenes and motions were captured, producing 12 blurry and 11 sharp video clips, each between 30 and 40 seconds long.

In deblurring PhoneCraft images, we used well-known blur datasets GoPro, REDS, and RSBlur. Results in \cref{fig:qual_phonecraft} show Blur2Blur significantly improved image clarity over pre-trained models, especially with RSBlur's complex blur patterns. The NIQE \cite{mittal2012making} scores of the deblurred images transformed by  Blur2Blur, using the GoPro, REDS, and RSBlur as source datasets are 9.8, 9.2, and 8.8, respectively. For NIQE score, lower is better, and this demonstrates Blur2Blur's ability to handle real-world blurs effectively.

\begin{figure}[t]
    \centering
    \includegraphics[width=0.75\linewidth]{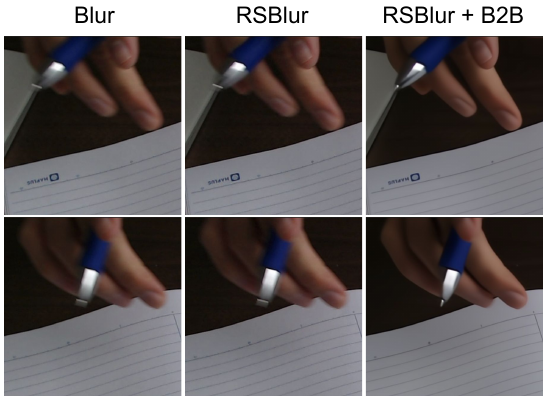}
    \vskip -0.1in 
    \caption{Results of using Blur2Blur on the WritingHands dataset. \label{fig:qual_hand}}
    \vspace{-5mm}
\end{figure}

In our second scenario, we explored a webcam-based application for monitoring hand movements during writing exercises, aimed at assisting in rehabilitation therapy. The challenge here is motion blur, which complicates hand and object tracking. To test our approach, we created a dataset named \textbf{WritingHands} with four 30fps webcam-recorded videos, each about 40s long. From these, two videos provided over 1100 frames with motion blur for training, and one video offered sharp reference images. Leveraging insights from the PhoneCraft dataset, we used the RSBlur dataset and its pre-trained NAFNet model for a two-day training session. Results, shown in \cref{fig:qual_hand}, indicate that while RSBlur's model alone leaves some blur, integrating it with Blur2Blur significantly restores the image's sharpness.

\section{Conclusions}
We have proposed Blur2Blur,  an effective approach to address the practical challenge of adapting image deblurring techniques to handle unseen blur. The key is to learn to convert an unknown blur to a known blur that can be effectively deblurred using a deblurring network specifically trained to handle the known blur. 
Throughout extensive experiments on synthetic and real-world benchmarks, Blur2Blur consistently exhibited superior performance, delivering impressive quantitative and qualitative outcomes. 

{
    \small
    \setlength{\bibsep}{0pt}
    \bibliographystyle{ieeenat_fullname}
    \bibliography{longstrings,main}

\begin{thebibliography}{46}
\providecommand{\natexlab}[1]{#1}
\providecommand{\url}[1]{\texttt{#1}}
\expandafter\ifx\csname urlstyle\endcsname\relax
  \providecommand{\doi}[1]{doi: #1}\else
  \providecommand{\doi}{doi: \begingroup \urlstyle{rm}\Url}\fi

\bibitem[Chan and Wong(1998)]{chan1998total}
Tony~F Chan and Chiu-Kwong Wong.
\newblock Total variation blind deconvolution.
\newblock \emph{IEEE Transactions on Image Processing}, 7\penalty0 (3):\penalty0 370--375, 1998.

\bibitem[Chen et~al.(2021)Chen, Lu, Zhang, Chu, and Chen]{chen2021hinet}
Liangyu Chen, Xin Lu, Jie Zhang, Xiaojie Chu, and Chengpeng Chen.
\newblock Hinet: Half instance normalization network for image restoration.
\newblock In \emph{Proceedings of the {IEEE} Conference on Computer Vision and Pattern Recognition}, 2021.

\bibitem[Chen et~al.(2022)Chen, Chu, Zhang, and Sun]{chen2022simple}
Liangyu Chen, Xiaojie Chu, Xiangyu Zhang, and Jian Sun.
\newblock Simple baselines for image restoration.
\newblock In \emph{Proceedings of the European Conference on Computer Vision}, 2022.

\bibitem[Cho et~al.(2021)Cho, Ji, Hong, Jung, and Ko]{mimounet}
Sung-Jin Cho, Seo-Won Ji, Jun-Pyo Hong, Seung-Won Jung, and Sung-Jea Ko.
\newblock Rethinking coarse-to-fine approach in single image deblurring.
\newblock In \emph{Proceedings of the International Conference on Computer Vision}, 2021.

\bibitem[Goodfellow et~al.(2020)Goodfellow, Pouget-Abadie, Mirza, Xu, Warde-Farley, Ozair, Courville, and Bengio]{goodfellow2020generative}
Ian Goodfellow, Jean Pouget-Abadie, Mehdi Mirza, Bing Xu, David Warde-Farley, Sherjil Ozair, Aaron Courville, and Yoshua Bengio.
\newblock Generative adversarial networks.
\newblock \emph{Communications of the ACM}, 63\penalty0 (11):\penalty0 139--144, 2020.

\bibitem[Gulrajani et~al.(2017)Gulrajani, Ahmed, Arjovsky, Dumoulin, and Courville]{gulrajani2017improved}
Ishaan Gulrajani, Faruk Ahmed, Martin Arjovsky, Vincent Dumoulin, and Aaron~C Courville.
\newblock Improved training of wasserstein gans.
\newblock \emph{Advances in Neural Information Processing Systems}, 30, 2017.

\bibitem[He et~al.(2016)He, Zhang, Ren, and Sun]{he2016deep}
Kaiming He, Xiangyu Zhang, Shaoqing Ren, and Jian Sun.
\newblock Deep residual learning for image recognition.
\newblock In \emph{Proceedings of the {IEEE} Conference on Computer Vision and Pattern Recognition}, 2016.

\bibitem[Hradi{\v{s}} et~al.(2015)Hradi{\v{s}}, Kotera, Zemc{\i}k, and {\v{S}}roubek]{hradivs2015convolutional}
Michal Hradi{\v{s}}, Jan Kotera, Pavel Zemc{\i}k, and Filip {\v{S}}roubek.
\newblock Convolutional neural networks for direct text deblurring.
\newblock In \emph{Proceedings of the British Machine Vision Conference}, 2015.

\bibitem[Isola et~al.(2017)Isola, Zhu, Zhou, and Efros]{isola2017image}
Phillip Isola, Jun-Yan Zhu, Tinghui Zhou, and Alexei~A Efros.
\newblock Image-to-image translation with conditional adversarial networks.
\newblock In \emph{Proceedings of the {IEEE} Conference on Computer Vision and Pattern Recognition}, 2017.

\bibitem[Johnson et~al.(2016)Johnson, Alahi, and Fei-Fei]{johnson2016perceptual}
Justin Johnson, Alexandre Alahi, and Li Fei-Fei.
\newblock Perceptual losses for real-time style transfer and super-resolution.
\newblock In \emph{Proceedings of the European Conference on Computer Vision}, 2016.

\bibitem[Jung et~al.(2020)Jung, Wada, Crall, Tanaka, Graving, Reinders, Yadav, Banerjee, Vecsei, Kraft, Rui, Borovec, Vallentin, Zhydenko, Pfeiffer, Cook, Fernández, De~Rainville, Weng, Ayala-Acevedo, Meudec, Laporte, et~al.]{imgaug}
Alexander~B. Jung, Kentaro Wada, Jon Crall, Satoshi Tanaka, Jake Graving, Christoph Reinders, Sarthak Yadav, Joy Banerjee, Gábor Vecsei, Adam Kraft, Zheng Rui, Jirka Borovec, Christian Vallentin, Semen Zhydenko, Kilian Pfeiffer, Ben Cook, Ismael Fernández, François-Michel De~Rainville, Chi-Hung Weng, Abner Ayala-Acevedo, Raphael Meudec, Matias Laporte, et~al.
\newblock {imgaug}.
\newblock \url{https://github.com/aleju/imgaug}, 2020.
\newblock Online; accessed 01-Feb-2020.

\bibitem[Kingma and Ba(2014)]{Kingma2014AdamAM}
Diederik~P. Kingma and Jimmy Ba.
\newblock Adam: A method for stochastic optimization.
\newblock \emph{CoRR}, abs/1412.6980, 2014.

\bibitem[Krishnan and Fergus(2009)]{krishnan2009fast}
Dilip Krishnan and Rob Fergus.
\newblock Fast image deconvolution using hyper-laplacian priors.
\newblock \emph{Advances in Neural Information Processing Systems}, 22:\penalty0 1033--1041, 2009.

\bibitem[Krishnan et~al.(2011)Krishnan, Tay, and Fergus]{krishnan2011blind}
Dilip Krishnan, Terence Tay, and Rob Fergus.
\newblock Blind deconvolution using a normalized sparsity measure.
\newblock In \emph{Proceedings of the {IEEE} Conference on Computer Vision and Pattern Recognition}. IEEE, 2011.

\bibitem[Kupyn et~al.(2018)Kupyn, Budzan, Mykhailych, Mishkin, and Matas]{kupyn2018deblurgan}
Orest Kupyn, Volodymyr Budzan, Mykola Mykhailych, Dmytro Mishkin, and Ji{\v{r}}{\'\i} Matas.
\newblock Deblurgan: Blind motion deblurring using conditional adversarial networks.
\newblock In \emph{Proceedings of the {IEEE} Conference on Computer Vision and Pattern Recognition}, 2018.

\bibitem[Kupyn et~al.(2019)Kupyn, Martyniuk, Wu, and Wang]{kupyn2019deblurgan}
Orest Kupyn, Tetiana Martyniuk, Junru Wu, and Zhangyang Wang.
\newblock Deblurgan-v2: Deblurring (orders-of-magnitude) faster and better.
\newblock In \emph{Proceedings of the International Conference on Computer Vision}, 2019.

\bibitem[Levin et~al.(2009)Levin, Weiss, Durand, and Freeman]{levin2009understanding}
Anat Levin, Yair Weiss, Fredo Durand, and William~T Freeman.
\newblock Understanding and evaluating blind deconvolution algorithms.
\newblock In \emph{Proceedings of the {IEEE} Conference on Computer Vision and Pattern Recognition}. IEEE, 2009.

\bibitem[Liang et~al.(2020)Liang, Chen, Liu, and Hsu]{liang2020raw}
Chih-Hung Liang, Yu-An Chen, Yueh-Cheng Liu, and Winston~H Hsu.
\newblock Raw image deblurring.
\newblock \emph{IEEE Transactions on Multimedia}, 24:\penalty0 61--72, 2020.

\bibitem[Liu et~al.(2014)Liu, Chang, and Ma]{liu2014blind}
Guangcan Liu, Shiyu Chang, and Yi Ma.
\newblock Blind image deblurring using spectral properties of convolution operators.
\newblock \emph{IEEE Transactions on Image Processing}, 23\penalty0 (12):\penalty0 5047--5056, 2014.

\bibitem[Lu et~al.(2019)Lu, Chen, and Chellappa]{lu2019unsupervised}
Boyu Lu, Jun-Cheng Chen, and Rama Chellappa.
\newblock Unsupervised domain-specific deblurring via disentangled representations.
\newblock In \emph{Proceedings of the {IEEE} Conference on Computer Vision and Pattern Recognition}, 2019.

\bibitem[Mehri et~al.(2021)Mehri, Ardakani, and Sappa]{mehri2021mprnet}
Armin Mehri, Parichehr~B Ardakani, and Angel~D Sappa.
\newblock Mprnet: Multi-path residual network for lightweight image super resolution.
\newblock In \emph{Proceedings of the IEEE/CVF Winter Conference on Applications of Computer Vision}, 2021.

\bibitem[Mittal et~al.(2012)Mittal, Soundararajan, and Bovik]{mittal2012making}
Anish Mittal, Rajiv Soundararajan, and Alan~C Bovik.
\newblock Making a “completely blind” image quality analyzer.
\newblock \emph{IEEE Signal processing letters}, 20\penalty0 (3):\penalty0 209--212, 2012.

\bibitem[Nah et~al.(2017{\natexlab{a}})Nah, Hyun~Kim, and Mu~Lee]{nah2017deep}
Seungjun Nah, Tae Hyun~Kim, and Kyoung Mu~Lee.
\newblock Deep multi-scale convolutional neural network for dynamic scene deblurring.
\newblock In \emph{Proceedings of the {IEEE} Conference on Computer Vision and Pattern Recognition}, 2017{\natexlab{a}}.

\bibitem[Nah et~al.(2017{\natexlab{b}})Nah, Kim, and Lee]{gopro}
Seungjun Nah, Tae~Hyun Kim, and Kyoung~Mu Lee.
\newblock Deep multi-scale convolutional neural network for dynamic scene deblurring.
\newblock In \emph{Proceedings of the {IEEE} Conference on Computer Vision and Pattern Recognition}, 2017{\natexlab{b}}.

\bibitem[Nah et~al.(2019)Nah, Timofte, Baik, Hong, Moon, Son, and Mu~Lee]{nah2019ntire}
Seungjun Nah, Radu Timofte, Sungyong Baik, Seokil Hong, Gyeongsik Moon, Sanghyun Son, and Kyoung Mu~Lee.
\newblock Ntire 2019 challenge on video deblurring: Methods and results.
\newblock In \emph{Proceedings of the IEEE/CVF Conference on Computer Vision and Pattern Recognition Workshops}, 2019.

\bibitem[Pham et~al.(2023)Pham, Tran, Tran, Pham, Nguyen, and Hoai]{pham2023hypercut}
Bang-Dang Pham, Phong Tran, Anh Tran, Cuong Pham, Rang Nguyen, and Minh Hoai.
\newblock Hypercut: Video sequence from a single blurry image using unsupervised ordering.
\newblock In \emph{Proceedings of the {IEEE} Conference on Computer Vision and Pattern Recognition}, 2023.

\bibitem[Ren et~al.(2020)Ren, Zhang, Wang, Hu, and Zuo]{ren2020neural}
Dongwei Ren, Kai Zhang, Qilong Wang, Qinghua Hu, and Wangmeng Zuo.
\newblock Neural blind deconvolution using deep priors.
\newblock In \emph{Proceedings of the {IEEE} Conference on Computer Vision and Pattern Recognition}, 2020.

\bibitem[Rim et~al.(2020)Rim, Lee, Won, and Cho]{rim2020real}
Jaesung Rim, Haeyun Lee, Jucheol Won, and Sunghyun Cho.
\newblock Real-world blur dataset for learning and benchmarking deblurring algorithms.
\newblock In \emph{Proceedings of the European Conference on Computer Vision}, 2020.

\bibitem[Rim et~al.(2022{\natexlab{a}})Rim, Kim, Kim, Lee, Lee, and Cho]{rim2022realistic}
Jaesung Rim, Geonung Kim, Jungeon Kim, Junyong Lee, Seungyong Lee, and Sunghyun Cho.
\newblock Realistic blur synthesis for learning image deblurring.
\newblock In \emph{Proceedings of the European Conference on Computer Vision}, 2022{\natexlab{a}}.

\bibitem[Rim et~al.(2022{\natexlab{b}})Rim, Kim, Kim, Lee, Lee, and Cho]{rsblur}
Jaesung Rim, Geonung Kim, Jungeon Kim, Junyong Lee, Seungyong Lee, and Sunghyun Cho.
\newblock Realistic blur synthesis for learning image deblurring.
\newblock In \emph{Proceedings of the European Conference on Computer Vision}, 2022{\natexlab{b}}.

\bibitem[Ronneberger et~al.(2015)Ronneberger, Fischer, and Brox]{ronneberger2015u}
Olaf Ronneberger, Philipp Fischer, and Thomas Brox.
\newblock U-net: Convolutional networks for biomedical image segmentation.
\newblock In \emph{Proceedings of the International Conference on Medical Image Computing and Computer Assisted Intervention}, 2015.

\bibitem[Shen et~al.(2019)Shen, Wang, Lu, Shen, Ling, Xu, and Shao]{shen2019human}
Ziyi Shen, Wenguan Wang, Xiankai Lu, Jianbing Shen, Haibin Ling, Tingfa Xu, and Ling Shao.
\newblock Human-aware motion deblurring.
\newblock In \emph{Proceedings of the International Conference on Computer Vision}, 2019.

\bibitem[Simonyan and Zisserman(2015)]{simonyan2015very}
K Simonyan and A Zisserman.
\newblock Very deep convolutional networks for large-scale image recognition.
\newblock In \emph{Proceedings of International Conference on Learning and Representation}, 2015.

\bibitem[Suin et~al.(2020)Suin, Purohit, and Rajagopalan]{suin2020spatially}
Maitreya Suin, Kuldeep Purohit, and AN Rajagopalan.
\newblock Spatially-attentive patch-hierarchical network for adaptive motion deblurring.
\newblock In \emph{Proceedings of the {IEEE} Conference on Computer Vision and Pattern Recognition}, 2020.

\bibitem[Tao et~al.(2018{\natexlab{a}})Tao, Gao, Shen, Wang, and Jia]{tao2018scale}
Xin Tao, Hongyun Gao, Xiaoyong Shen, Jue Wang, and Jiaya Jia.
\newblock Scale-recurrent network for deep image deblurring.
\newblock In \emph{Proceedings of the {IEEE} Conference on Computer Vision and Pattern Recognition}, 2018{\natexlab{a}}.

\bibitem[Tao et~al.(2018{\natexlab{b}})Tao, Gao, Shen, Wang, and Jia]{tao2018srndeblur}
Xin Tao, Hongyun Gao, Xiaoyong Shen, Jue Wang, and Jiaya Jia.
\newblock Scale-recurrent network for deep image deblurring.
\newblock In \emph{Proceedings of the {IEEE} Conference on Computer Vision and Pattern Recognition}, 2018{\natexlab{b}}.

\bibitem[Torbunov et~al.(2023)Torbunov, Huang, Tseng, Yu, Huang, Yoo, Lin, Viren, and Ren]{torbunov2023rethinking}
Dmitrii Torbunov, Yi Huang, Huan-Hsin Tseng, Haiwang Yu, Jin Huang, Shinjae Yoo, Meifeng Lin, Brett Viren, and Yihui Ren.
\newblock Rethinking cyclegan: Improving quality of gans for unpaired image-to-image translation.
\newblock \emph{arXiv preprint arXiv:2303.16280}, 2023.

\bibitem[Tran et~al.(2021)Tran, Tran, Phung, and Hoai]{tran2021explore}
Phong Tran, Anh~Tuan Tran, Quynh Phung, and Minh Hoai.
\newblock Explore image deblurring via encoded blur kernel space.
\newblock In \emph{Proceedings of the {IEEE} Conference on Computer Vision and Pattern Recognition}, 2021.

\bibitem[Vakunov et~al.(2020)Vakunov, Chang, Zhang, Sung, Grundmann, and Bazarevsky]{vakunov2020mediapipe}
Andrey Vakunov, Chuo-Ling Chang, Fan Zhang, George Sung, Matthias Grundmann, and Valentin Bazarevsky.
\newblock Mediapipe hands: On-device real-time hand tracking.
\newblock In \emph{Proceedings of the IEEE/CVF Conference on Computer Vision and Pattern Recognition Workshops}, 2020.

\bibitem[Yi et~al.(2017)Yi, Zhang, Tan, and Gong]{yi2017dualgan}
Zili Yi, Hao Zhang, Ping Tan, and Minglun Gong.
\newblock Dualgan: Unsupervised dual learning for image-to-image translation.
\newblock In \emph{Proceedings of the International Conference on Computer Vision}, 2017.

\bibitem[Zamir et~al.(2021)Zamir, Arora, Khan, Hayat, Khan, Yang, and Shao]{zamir2021multi}
Syed~Waqas Zamir, Aditya Arora, Salman Khan, Munawar Hayat, Fahad~Shahbaz Khan, Ming-Hsuan Yang, and Ling Shao.
\newblock Multi-stage progressive image restoration.
\newblock In \emph{Proceedings of the {IEEE} Conference on Computer Vision and Pattern Recognition}, 2021.

\bibitem[Zamir et~al.(2022)Zamir, Arora, Khan, Hayat, Khan, and Yang]{zamir2022restormer}
Syed~Waqas Zamir, Aditya Arora, Salman Khan, Munawar Hayat, Fahad~Shahbaz Khan, and Ming-Hsuan Yang.
\newblock Restormer: Efficient transformer for high-resolution image restoration.
\newblock In \emph{Proceedings of the {IEEE} Conference on Computer Vision and Pattern Recognition}, 2022.

\bibitem[Zhang et~al.(2021)Zhang, Liang, Van~Gool, and Timofte]{zhang2021designing}
Kai Zhang, Jingyun Liang, Luc Van~Gool, and Radu Timofte.
\newblock Designing a practical degradation model for deep blind image super-resolution.
\newblock In \emph{Proceedings of the International Conference on Computer Vision}, 2021.

\bibitem[Zhao et~al.(2022)Zhao, Zhang, Hong, Xu, Yang, and Wang]{zhao2022fcl}
Suiyi Zhao, Zhao Zhang, Richang Hong, Mingliang Xu, Yi Yang, and Meng Wang.
\newblock Fcl-gan: A lightweight and real-time baseline for unsupervised blind image deblurring.
\newblock In \emph{Proceedings of the 30th ACM International Conference on Multimedia}, 2022.

\bibitem[Zhong et~al.(2020)Zhong, Gao, Zheng, and Zheng]{zhong2020efficient}
Zhihang Zhong, Ye Gao, Yinqiang Zheng, and Bo Zheng.
\newblock Efficient spatio-temporal recurrent neural network for video deblurring.
\newblock In \emph{Proceedings of the European Conference on Computer Vision}, 2020.

\bibitem[Zhu et~al.(2017)Zhu, Park, Isola, and Efros]{zhu2017unpaired}
Jun-Yan Zhu, Taesung Park, Phillip Isola, and Alexei~A Efros.
\newblock Unpaired image-to-image translation using cycle-consistent adversarial networks.
\newblock In \emph{Proceedings of the International Conference on Computer Vision}, 2017.

\end{thebibliography}
}

\clearpage
\setcounter{page}{1}
\maketitlesupplementary

\begin{abstract}
   In this supplementary PDF, we first provide the qualitative results obtained by methods with the Restormer backbone \cite{zamir2022restormer} and some additional qualitative results of each dataset to show our effectiveness in deblurring unknown-blur images compared to other baselines. Next, we illustrate the performance with different backbones for the Blur2Blur translator model. Finally, we provide details of our collected PhoneCraft dataset and validate the video deblurring performance of Blur2Blur, demonstrating significant enhancements in hand movement visualization and thus leaving room for practical application. We also include our code and a video of sample deblurring results in the supplementary package.
\end{abstract}

\section{Additional Qualitative Results}
\subsection{Restormer model}
\begin{figure*}[!htb]
    \centering
    \includegraphics[width=\textwidth]{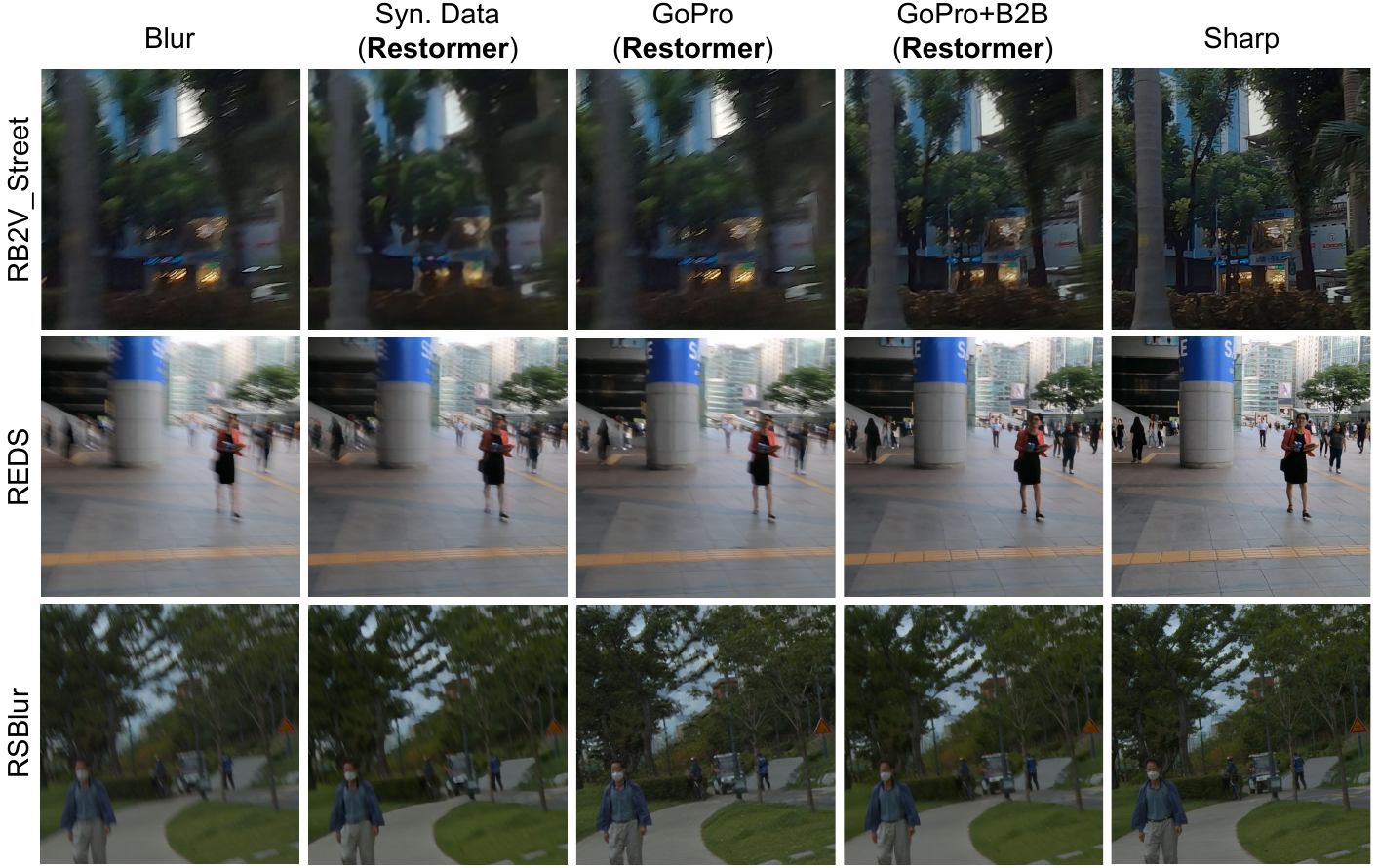}
    \caption{Qualitative results of Restormer \cite{zamir2022restormer} on three datasets.}
    \label{fig:restormer}
\end{figure*}

In \cref{fig:qual1} in the main paper, we omit the results with the Restormer backbone due to the space limit. We provide these results in this supplementary in \cref{fig:restormer}. As can be seen, Restormer shows behavior similar to NAFNet. The original network produces blurry images that are close to the input images. However, when combined with Blur2Blur, it can successfully deblur the images and produce sharper outputs. From quantitative numbers, Restormer-based models perform slightly worse than the NAFNet-based counterparts.

\subsection{Additional Deblurring Results}
\begin{figure*}[t]
    \centering
    \includegraphics[width=0.95\textwidth]{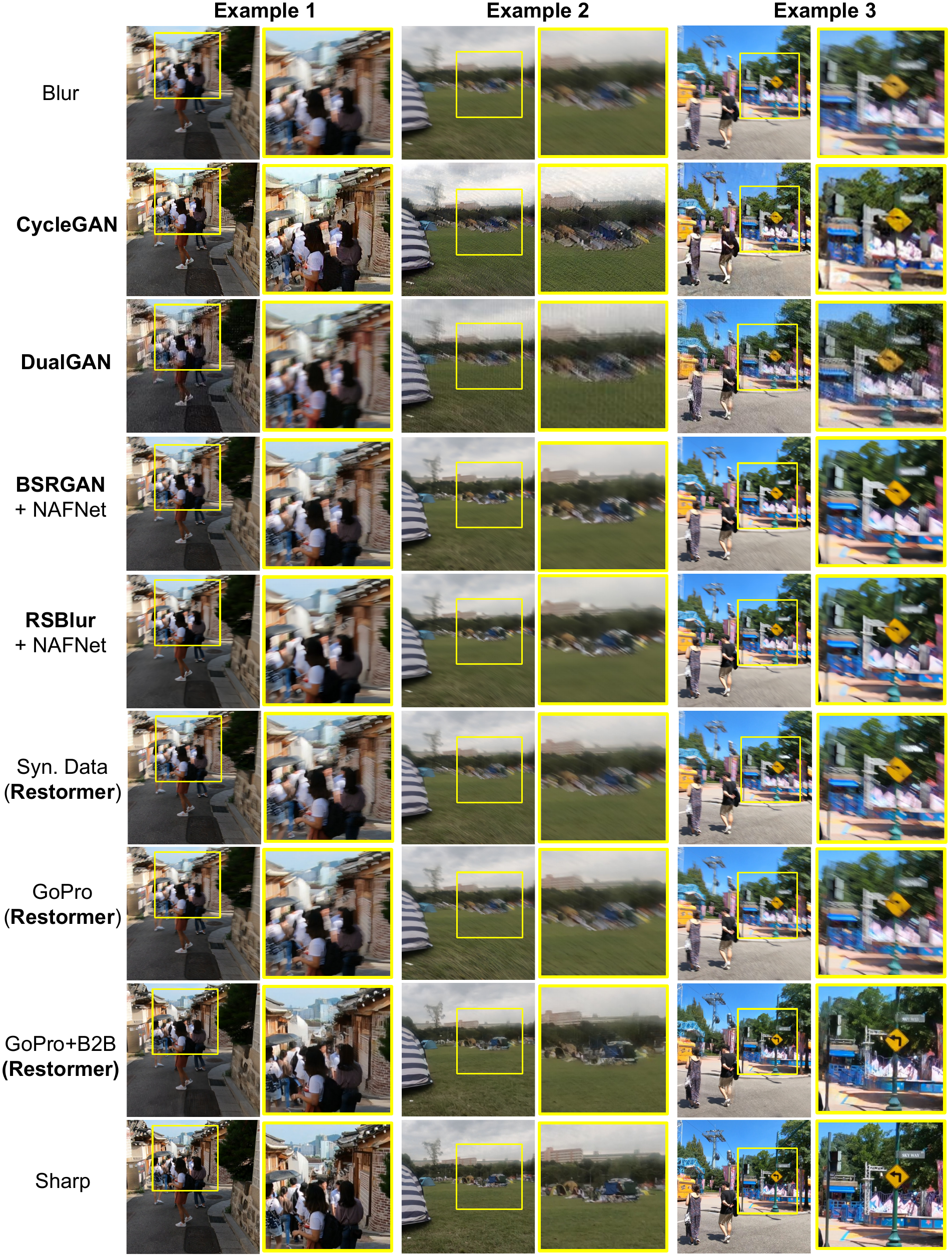}
    \caption{Extra qualitative results on the REDS dataset.}
    \label{fig:exp2_reds}
\end{figure*}

\begin{figure*}[t]
    \centering
    \includegraphics[width=0.95\textwidth]{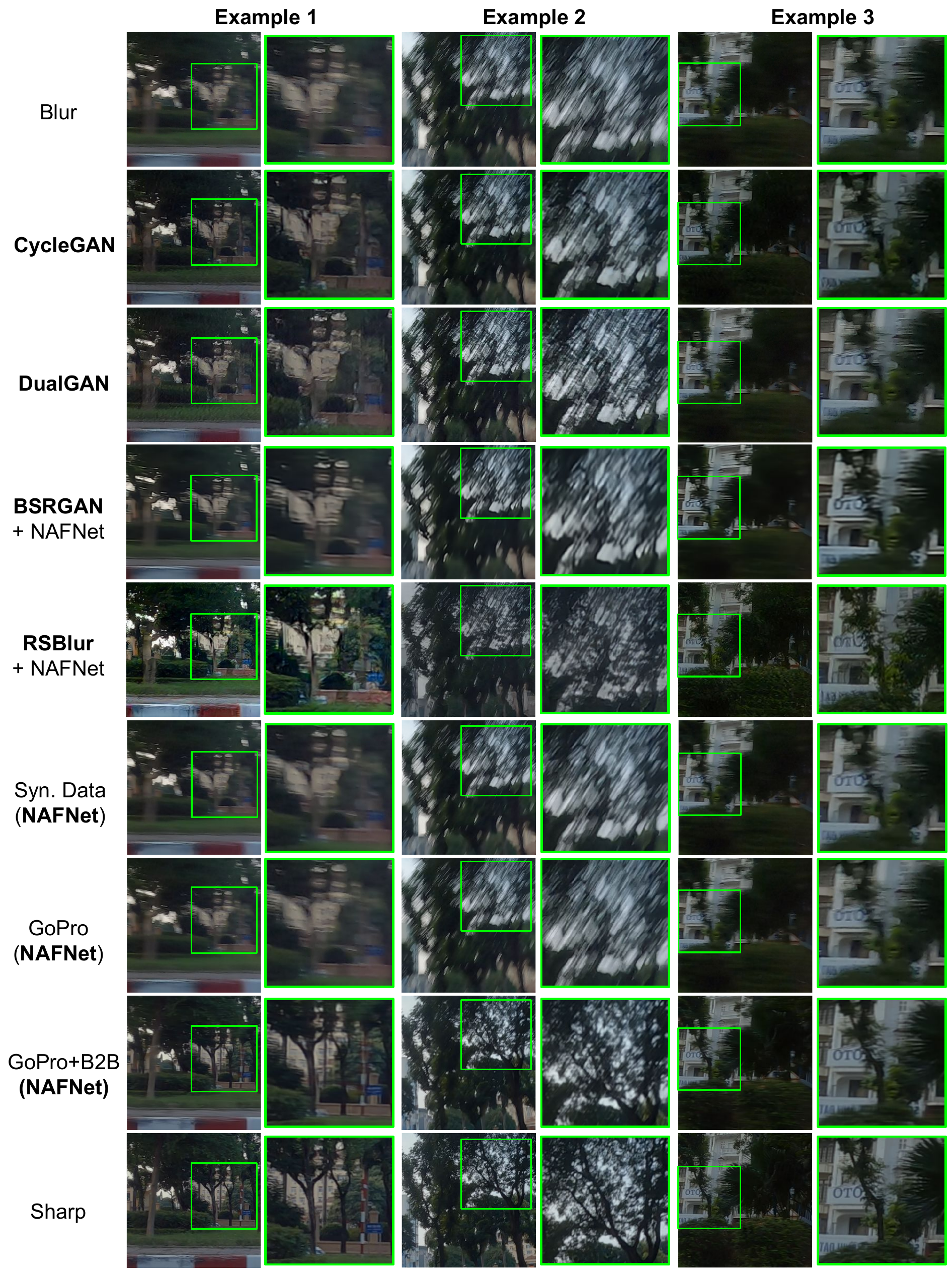}
    \caption{Extra qualitative results on the RB2V\_Street dataset.}
    \label{fig:exp2_rb2v}
\end{figure*}

\begin{figure*}[t]
    \centering
    \includegraphics[width=\textwidth]{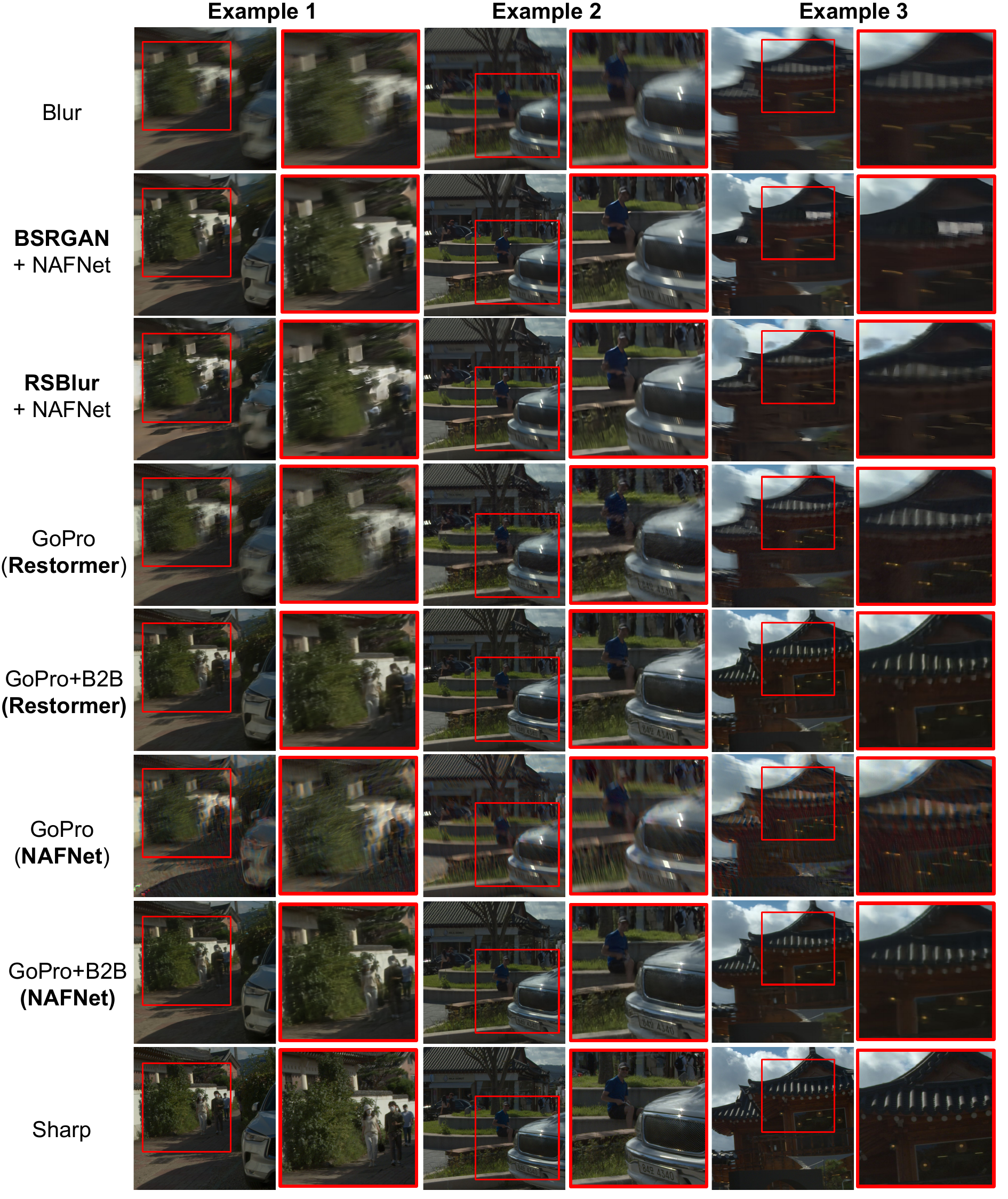}
    \caption{Extra qualitative results on the RSBlur dataset.}
    \label{fig:exp2_rsblur}
\end{figure*}

In this section, we provide additional qualitative figures comparing the image deblurring results of our Blur2Blur and other baselines. Figures \ref{fig:exp2_reds}, \ref{fig:exp2_rb2v}, and \ref{fig:exp2_rsblur} show samples where $\mK$ is built upon the GoPro dataset \cite{gopro}, with the Unknown set derived respectively from the REDS dataset \cite{nah2019ntire}, RB2V\_Street \cite{pham2023hypercut}, and RSBlur \cite{rsblur}.

\section{Blur2Blur Analysis}
\subsection{Backbone Experiments}
We explore the integration of multi-scale architectures into the Blur2Blur mechanism by experimenting with different backbones. The UNet architecture \cite{ronneberger2015u} has been adapted to handle inputs at various scales, allowing for a more nuanced understanding of blur at multiple scales. Concurrently, we employed the NAFNet backbone in its original form, taking advantage of its robust feature extraction capabilities without modifications. The result on \cref{tab:backbone} shows that MIMO-UNet clearly surpasses the standard UNet, even in its modified form. Moreover, the results also reveal that the NAFNet does not perform as well as the multi-scale variants, highlighting the importance of multi-scale level optimization in the Blur2Blur framework for deblurring tasks.
\begin{table}[!htb] {
    \centering
    \begin{tabu}{lcc}
        \toprule
        Backbone & PSNR$\uparrow$ & SSIM$\uparrow$ \\
        \midrule
        UNet \cite{ronneberger2015u}  & 22.54 & 0.732 \\
        MIMO-UNet \cite{mimounet} & 26.98 & 0.812 \\ 
        NAFNet \cite{chen2022simple} & 20.54 & 0.686 \\        
        \bottomrule
    \end{tabu}    
    \caption{Ablation studies with the Blur2Blur backbone. \label{tab:backbone}}
}
\end{table}

\subsection{Impact of Sataset Size.} 
In \cref{tab:psnr}, we validate the deblurring performance using GoPro-RB2V datasets, maintaining a fixed $\mB$:$\mS$ ratio of 6:4 while varying the dataset size across four different scales of the target dataset ($\alpha$).
\begin{table}[!htb]
        \centering
       
        \begin{tabular}{ccccc}
            \toprule
            $\alpha$ & 0.25 & 0.5 & 0.75 & 1.0 \\
            \midrule
            PSNR & 25.45 & 25.93 & 26.32 & 26.98 \\
            \bottomrule
        \end{tabular}
        \caption{Affect of data size}
         \label{tab:psnr}
    \end{table}%

\subsection{Validation on the blur converter} 
To evaluate the effectiveness of our blur converter, we do two classification experiments to determine the alignment of converted images with the KnownBlur domain ($\mK$), using GoPro-RB2V settings as detailed in the main paper. The first (\textbf{Acc1}) used the pretrained Discriminator from our Blur2Blur framework, assessing if converted images belongs to ($\mK$). For the second (\textbf{Acc2}), we synthesized a new dataset via the Blur Kernel Extractor \textit{F} \cite{tran2021explore} using sharp GoPro images combined with blur kernels from the target datasets. We then trained a ResNet18 \cite{he2016deep} to discern if the blur in converted images corresponded to ($\mK$) or not. In both experiments, our method converts the input image to have the target known blur with near 100\% accuracy (\cref{tab:3}). 


\begin{table}[!htb]
    \centering
    \begin{tabular}{c|cc}
        \toprule
        Model & Input & Converted \\
        \midrule
        Acc1/\textcolor{teal}{Acc2}(\%) & 11.24 / \textcolor{teal}{6.85} & 86.67 / \textcolor{teal}{96.53} \\
        \bottomrule
    \end{tabular}
    \caption{Blur converter validation}
    \label{tab:3}
\end{table}
    
\section{Real-world Application}
\subsection{Details of PhoneCraft collection}
The data collection process for training Blur2Blur is actually inexpensive. Although the number of images required looks high (several thousand for each subset), they are mostly video frames and thus can be collected effectively. For example, in the PhoneCraft experiment above, we only need to collect 11 sharp videos and 12 blurry ones, with a total collection time of less than 2 hours. More specifically, the dataset contains more than 12500 diverse blurry images and 11000 sharp images. 

\subsection{Video Deblurring Performance}
As mentioned in the main paper, to enrich our practical evaluation with more tangible visual examples and to demonstrate one real-world application of our Blur2Blur mode, we incorporated a video from the collected dataset. This video simulates scenarios with significant motion blur, which is common in dynamic environments. The clarity of visual details in such situations is crucial for various applications, including rehabilitation therapy. Accurate hand movement visualization is vital for tasks like hand pose detection and gesture-based interactive rehabilitation systems.

To evaluate our Blur2Blur model, we used a video with pronounced hand movements, pre-training the deblurring model on the RSBlur dataset. The results, demonstrated in \red{video1.mp4}, clearly show that our Blur2Blur framework significantly enhances visual clarity compared to using the pre-trained deblurring model alone. Moreover, to further assess the enhancement in hand movement recognition, we validated the deblurred videos using the Hand Pose Estimation model from MediaPipe\cite{vakunov2020mediapipe}. The results, shown in the video, highlight a notable improvement in hand pose estimation when using our method. The enhanced sharpness and detail achieved by Blur2Blur enable more accurate and reliable recognition of hand poses. This demonstrates the potential of our Blur2Blur model in applications demanding high-fidelity visualization of hand movements, especially in advanced rehabilitation therapy tools that rely on precise hand movement tracking for effective patient care and recovery.

Besides that, we also provide the additional qualitative video deblurring result in PhoneCraft dataset is illustrated in \red{video2.mp4}.



\end{document}